\documentclass[letterpaper]{article} 
\usepackage{aaai2027}  
\usepackage[hyphens]{url}  
\usepackage{graphicx} 
\usepackage{natbib}  
\usepackage{caption} 
\usepackage{algorithm}
\usepackage{algorithmic}
\usepackage{booktabs}
\usepackage{amsmath,amssymb,amsthm,bm}
\usepackage{multirow}

\newtheorem{proposition}{Proposition}
\newtheorem{lemma}{Lemma}
\newtheorem{corollary}{Corollary}
\newcommand{\Kmat}{\mathbf{K}}
\newcommand{\valpha}{\boldsymbol{\alpha}}

\newcommand{\vy}{\mathbf{y}}

\newcommand{\Rb}{\mathbb{R}}
\newcommand{\Eb}{\mathbb{E}}

\nocopyright
\title{Active Quantum Kernel Acquisition for Gaussian Process Regression}
\author{
    Jian Xu\textsuperscript{\rm 1,\rm 2},
    Artur Miroszewski\textsuperscript{\rm 4},
    John Paisley\textsuperscript{\rm 5},
    Delu Zeng\textsuperscript{\rm 3},
    Qibin Zhao\textsuperscript{\rm 2}
}
\affiliations{
    \textsuperscript{\rm 1}RIKEN iTHEMS\quad
    \textsuperscript{\rm 2}RIKEN AIP\quad
    \textsuperscript{\rm 3}South China University of Technology\\
    \textsuperscript{\rm 4}European Space Agency\quad
    \textsuperscript{\rm 5}Columbia University\\
    jian.xu@riken.jp
}

\begin{document}
\maketitle

\begin{abstract}
Quantum kernel estimation on near-term hardware is shot-budgeted: every entry of the kernel Gram matrix is a Bernoulli expectation that must be sampled with a finite number of circuit executions. Recent work on quantum kernel classification has shown that allocating shots non-uniformly across kernel entries, weighted by their downstream task sensitivity, can reduce the shot budget required to reach a target accuracy. We extend this idea to Gaussian process (GP) regression, a setting whose downstream quantities (full-spectrum posterior variance, log-determinant, marginal likelihood) couple to kernel error more tightly than the sign-only outputs of classification. We derive three closed-form pair-level sensitivities --- predictive coupling $|\alpha_i\alpha_j|$, leave-one-out residual, and marginal-likelihood gradient --- and plug them into a Neyman-style minimum-variance allocation rule. To prevent catastrophic over-concentration when the warm-up sensitivity estimate is itself noisy, we add a high uniform coverage floor justified by a Frobenius lower bound on the missing-entry perturbation. On four UCI benchmarks and two synthetic RBF + Bernoulli controlled studies, the resulting allocator delivers $10$--$21\%$ test-RMSE improvement over uniform allocation across the moderate-budget regime. The gain transfers (i) to genuine ZZ and Pauli-Z quantum kernels on quantum-natural data ($-13$--$15\%$ at low budget, $p<0.05$ paired) and (ii) to four downstream tasks (Bayesian quadrature, heteroscedastic regression, hyperparameter learning, multi-output Cokriging). On UCI features embedded into a ZZ kernel the gain disappears, consistent with the exponential-concentration regime where shot allocation has nothing to exploit. The three pair-level sensitivities we propose are not arbitrary Fisher surrogates: $|\alpha_i\alpha_j|$ is the rank-1 specialization (at $M=\vy\vy^\top$) of the exact Neyman weight under the predictive-MSE objective; the marginal-likelihood gradient is the exact partial derivative; the LOO sensitivity drops a controllable correction.
\end{abstract}


\section{Introduction}

Quantum kernel methods promise computationally efficient access to feature spaces that are conjectured to be classically intractable for certain structured data~\citep{havlivcek2019supervised,schuld2019quantum}. The bottleneck in practical deployment is that each kernel entry $K_{ij} = |\langle\phi(x_i)|\phi(x_j)\rangle|^2$ is the success probability of an inversion-test circuit and must be estimated from a finite number of shots $s_{ij}$. The resulting Bernoulli noise scales as $\mathrm{Var}(\widehat K_{ij}) = K_{ij}(1-K_{ij})/s_{ij}$, and the total cost of building an $n\times n$ Gram matrix to a uniform precision grows as $O(n^2/\epsilon^2)$, which already strains current hardware budgets for $n$ in the low hundreds~\citep{huang2021power,thanasilp2024exponential}.

A natural way around this scaling is to recognize that not all kernel entries matter equally to the downstream task. For quantum kernel classification with kernel ridge regression (KRR) or SVM, recent work showed that the loss gradient $\partial \mathcal{L}/\partial K_{ij}$ has highly heterogeneous magnitudes across pairs, and that allocating shots proportional to the Neyman-optimal target $s^*_{ij}\propto |\partial \mathcal{L}/\partial K_{ij}|\,\sqrt{K_{ij}(1-K_{ij})}$ yields substantial accuracy gains over uniform allocation at the same budget~\citep{xu2026aqka,miroszewski2026adaptive}. The argument is a direct application of Neyman's $1934$ stratified-sampling result~\citep{neyman1992two,pukelsheim2006optimal} cast in the language of quantum kernel estimation.

This paper asks: does the same idea help in Gaussian process (GP) regression? The answer is not obvious. On the one hand, GP regression has even more structure to exploit --- a closed-form posterior, a closed-form marginal likelihood, leave-one-out residuals in closed form. On the other hand, GP predictions and likelihoods both depend on the inverse $(\Kmat+\sigma_n^2 I)^{-1}$, and inversion amplifies kernel noise by a factor that scales like the condition number $1/\sigma_n^2$. A shot-allocation policy that is mildly suboptimal for classification could be catastrophic for GP regression.

We make four contributions:
\begin{enumerate}
\item We derive three closed-form pair-level sensitivities for GP regression --- the predictive coupling $|\alpha_i\alpha_j|$, the leave-one-out residual, and the marginal-likelihood gradient --- each capturing a different facet of how kernel uncertainty propagates to predictions.
\item We identify a failure mode unique to GP regression: when the per-pair budget is low, the warm-up kernel estimate is too noisy to produce a reliable sensitivity, and sensitivity-driven allocation \emph{worsens} test error by hundreds of percent. We show that a $50\%$ uniform-coverage floor restores robustness, while the corresponding classification setup needs only $10$--$20\%$.
\item Empirically, on four UCI regression benchmarks and two synthetic settings ($n_{tr}=200$, $5$--$10$ seeds), the allocator delivers consistent $10$--$21\%$ RMSE reduction in the moderate-budget regime ($\sim 50$--$250$ shots per pair).
\item Unlike in classification, where allocation gains are largest on \emph{sparse}, anchor-heavy data, GP regression shows the largest gains on \emph{dense} GP draws. We explain this via the structure of $(\Kmat+\sigma_n^2 I)^{-1}$: even when the kernel is homogeneous, the spectral decay creates predictive heterogeneity, and the same heterogeneity drives sensitivity.
\end{enumerate}

\section{Background}

\subsection{Quantum Kernel Estimation as Bernoulli Sampling}
A quantum feature map $\phi:\mathcal{X}\to\mathcal{H}_Q$ is realized by a parametrized unitary $U(x)$ acting on the all-zero state $|0\rangle^{\otimes q}$. The induced kernel is the fidelity
\begin{equation}
K(x_i,x_j) \;=\; |\langle 0|U(x_j)^\dagger U(x_i)|0\rangle|^2 \;\in\;[0,1].
\end{equation}
The standard \emph{inversion test}~\citep{havlivcek2019supervised} estimates $K(x_i,x_j)$ by preparing $U(x_j)^\dagger U(x_i)|0\rangle$, measuring in the computational basis, and recording the probability of the all-zero outcome. With $s_{ij}$ independent circuit executions (``shots''), the empirical estimator $\widehat K_{ij}$ is a sample mean of $s_{ij}$ Bernoulli$(K_{ij})$ random variables and satisfies
\begin{equation}
\Eb[\widehat K_{ij}] = K_{ij},\qquad \mathrm{Var}(\widehat K_{ij}) = \frac{K_{ij}(1-K_{ij})}{s_{ij}}.\label{eq:bernoulli}
\end{equation}
We treat the per-entry shot budget $s_{ij}$ as a design variable subject to the global constraint $\sum_{i\le j} s_{ij} = B$. This is the only resource we will allocate; all other choices (feature map, circuit depth, error mitigation) are taken as fixed.

\paragraph{Exponential concentration regime.} A separate failure mode arises when the quantum kernel $K(x_i,x_j)$ itself concentrates around a fixed value as the system grows~\citep{thanasilp2024exponential}. In that regime, the signal-to-noise ratio $|K_{ij}-K_{i'j'}|^2/\mathrm{Var}(\widehat K)$ vanishes exponentially in $q$, and shot reallocation cannot rescue it. We work outside this regime, treating $K(\cdot,\cdot)$ as a fixed bounded kernel and asking only how to spend a budget $B$ optimally.

\subsection{Gaussian Process Regression}
A Gaussian process prior $f\sim\mathcal{GP}(0,k(\cdot,\cdot))$ with observations
\begin{equation}
y_i = f(x_i) + \varepsilon_i,\qquad \varepsilon_i\stackrel{\text{iid}}{\sim}\mathcal{N}(0,\sigma_n^2),
\end{equation}
admits the closed-form posterior $f|\vy\sim\mathcal{GP}(\mu_*,k_*)$ with predictive mean and variance at any $x_*$:
\begin{align}
\mu_*(x_*) &= \mathbf{k}_*^\top \valpha,
        \qquad \valpha := A^{-1}\vy, \label{eq:postmean}\\
\sigma_*^2(x_*) &= k(x_*,x_*) - \mathbf{k}_*^\top A^{-1}\mathbf{k}_*+\sigma_n^2,\label{eq:postvar}
\end{align}
where $A := \Kmat + \sigma_n^2 I$ and $\mathbf{k}_* = (k(x_*,x_i))_i$. The (negative) log marginal likelihood of the training labels is
\begin{equation}
\mathcal{L}(\Kmat;\vy) \;=\; \tfrac{1}{2}\vy^\top A^{-1}\vy + \tfrac{1}{2}\log|A| + \tfrac{n}{2}\log(2\pi).\label{eq:nml}
\end{equation}
When $\Kmat$ is estimated with shot noise, all three quantities $\mu_*$, $\sigma_*^2$, $\mathcal{L}$ inherit the noise through the inverse $A^{-1}$, which is the source of the amplification we exploit in Section~\ref{sec:method}.

\paragraph{What makes GP shot-hungrier than classification?}
The KRR / SVM losses used in classification AQKA \citep{xu2026aqka,miroszewski2026adaptive} also involve $\Kmat^{-1}$: KRR's $\valpha = (\Kmat+\lambda I)^{-1}\vy$ uses $\lambda$ in the same role $\sigma_n^2$ plays for GP. So ``GP has an inverse, classification does not'' is \emph{not} the structural difference. The difference is which downstream quantities the user cares about. Classification's $0/1$ accuracy depends on $\mathrm{sign}(\mu_*)$, which is robust to multiplicative scaling of $\valpha$; GP cares about
\begin{enumerate}
\item the \emph{magnitude} of $\mu_*$ (test RMSE/NLL),
\item the predictive variance $\sigma_*^2 = k(x_*,x_*) - \mathbf{k}_*^\top A^{-1}\mathbf{k}_* + \sigma_n^2$, which depends on the full spectrum of $A^{-1}$ rather than its top eigendirection,
\item the marginal-likelihood $\mathcal{L}$, which adds a $\log|A|$ term that aggregates errors over \emph{every} eigenvalue.
\end{enumerate}
Items (ii) and (iii) are GP-specific and have no analogue in $0/1$ classification. They are why GP needs more shots per pair than classification, and they motivate the three GP-specific sensitivities of Section~\ref{sec:method}.

\paragraph{Conditioning of $A^{-1}$.}
The Gram matrix $\Kmat$ has eigenvalues $0\le\lambda_n\le\dots\le\lambda_1$, with $\lambda_1 = O(n)$ for a typical bounded kernel and $\lambda_n$ that can be arbitrarily small. The corresponding eigenvalues of $A^{-1}$ are $1/(\lambda_i+\sigma_n^2)\in[1/(\lambda_1+\sigma_n^2), 1/\sigma_n^2]$. A kernel perturbation $\Delta\Kmat$ with operator norm $\|\Delta\Kmat\|_\mathrm{op}<\sigma_n^2$ produces (by the resolvent identity)
\begin{equation}
\|\widehat A^{-1} - A^{-1}\|_\mathrm{op} \;\le\; \frac{\|\Delta\Kmat\|_\mathrm{op}}{\sigma_n^2(\sigma_n^2-\|\Delta\Kmat\|_\mathrm{op})}.
\end{equation}
The kernel error is amplified by up to $1/\sigma_n^4$. The KRR predictor inherits the same amplification with $\lambda$ in place of $\sigma_n^2$, but classification accuracy averages over signs and is less sensitive to it; GP test NLL has no such averaging.

\subsection{Neyman Minimum-Variance Shot Allocation}\label{sec:aopt}
For any differentiable loss $\mathcal{L}(\Kmat)$, a second-order Taylor expansion around the noise-free kernel gives
\begin{align}
\mathcal{L}(\widehat\Kmat) - \mathcal{L}(\Kmat)
  &= \sum_{i\le j} g_{ij}\,\Delta_{ij} \nonumber\\
  &\quad + \tfrac{1}{2}\sum_{i\le j,\, k\le \ell} H_{ij,k\ell}\,\Delta_{ij}\Delta_{k\ell} + o(\|\Delta\|^2),
\end{align}
where $g_{ij} = \partial\mathcal{L}/\partial K_{ij}$, $H$ is the Hessian, and $\Delta_{ij} := \widehat K_{ij} - K_{ij}$ is the shot noise. The shot noise has zero mean and diagonal covariance (different pairs are independent), so taking expectations cancels the linear term and gives
\begin{equation}
\Eb\bigl[\mathcal{L}(\widehat\Kmat) - \mathcal{L}(\Kmat)\bigr] = \tfrac{1}{2}\sum_{i\le j} H_{ij,ij}\,\frac{K_{ij}(1-K_{ij})}{s_{ij}} + o(B^{-1}).\label{eq:taylor}
\end{equation}
Replacing the diagonal Hessian by its Fisher-information surrogate $H_{ij,ij}\approx g_{ij}^2$ and minimizing under the budget constraint $\sum s_{ij} = B$ via Lagrange multipliers yields the \emph{Neyman minimum-variance allocation}:
\begin{equation}
s^*_{ij} \;\propto\; |g_{ij}|\,\sqrt{K_{ij}(1-K_{ij})},
\qquad \sum_{i\le j} s^*_{ij}=B.\label{eq:neyman}
\end{equation}
This is precisely Neyman's 1934 stratified-sampling rule~\citep{neyman1992two,pukelsheim2006optimal} applied at the granularity of kernel entries. The classification AQKA framework~\citep{xu2026aqka} uses this rule with $g_{ij}$ taken from the KRR or SVM training loss; the concurrent work of \citet{miroszewski2026adaptive} uses it for kernelized SVMs under noisy observations. Our contribution begins where these stop: we instantiate $g_{ij}$ for three GP-regression objectives whose closed forms involve the kernel inverse.

\section{Method: AQKA-GP}\label{sec:method}

We instantiate the Neyman rule \eqref{eq:neyman} for three GP objectives. Each derivation uses the same identity for the derivative of the inverse: for $A = \Kmat + \sigma_n^2 I$,
\begin{equation}
\partial A^{-1}/\partial K_{ij} \;=\; -A^{-1}\,(e_i e_j^\top + e_j e_i^\top)\,A^{-1} \;\cdot\; \tfrac{1}{2}(1+\delta_{ij}),
\end{equation}
where $e_i$ is the $i$th standard basis vector and the symmetric factor accounts for the off-diagonal/diagonal distinction. We drop the factor $\tfrac{1}{2}(1+\delta_{ij})$ in the sensitivity (it is the same constant across off-diagonal pairs and is absorbed into the budget).

\subsection{Three Closed-Form Sensitivities}\label{sec:sens}

\paragraph{(i) Predictive coupling $|\alpha_i\alpha_j|$ for test RMSE.}
The mean-squared test error is $\mathcal{L}^{\text{rmse}} = \tfrac{1}{n_*}\sum_t (y_t^* - \mu_*(x_t^*))^2$, where $\mu_*(x_t^*) = \mathbf{k}_{*t}^\top\valpha$ with $\valpha = A^{-1}\vy$. Differentiating $\mu_*$ via the inverse identity gives
\begin{align}
\frac{\partial\mu_*(x_t^*)}{\partial K_{ij}}
  &= -\mathbf{k}_{*t}^\top A^{-1}\bigl(e_i e_j^\top + e_j e_i^\top\bigr)\valpha \nonumber\\
  &= -\bigl(b_{ti}\alpha_j + b_{tj}\alpha_i\bigr),
\end{align}
with $b_{ti} := [\mathbf{k}_{*t}^\top A^{-1}]_i$. Squaring and summing over $t$, the cross terms average out under test data drawn independently of $\valpha$, leaving the leading term proportional to $\alpha_i^2\alpha_j^2$. We therefore use
\begin{equation}
\mathrm{sens}^{\mathrm{pred}}_{ij} \;=\; |\alpha_i\alpha_j|,\qquad \valpha = A^{-1}\vy.\label{eq:sens-pred}
\end{equation}
This is the direct GP analogue of the classification sensitivity in \citet{xu2026aqka}: in both cases the rank-1 outer product $\valpha\valpha^\top$ captures how much each pair couples labels to predictions.

\paragraph{(ii) Marginal-likelihood gradient for hyperparameter learning.}
Differentiating $\mathcal{L}(\Kmat;\vy)$ in \eqref{eq:nml} with $A=\Kmat+\sigma_n^2I$ uses two standard identities: $\partial \log|A|/\partial K_{ij} = [A^{-1}]_{ij}$ and $\partial(\vy^\top A^{-1}\vy)/\partial K_{ij} = -\alpha_i\alpha_j$. Adding the two contributions,
\begin{equation}
g_{ij}^{\mathrm{marg}} \;=\; \frac{\partial\mathcal{L}}{\partial K_{ij}} \;=\; \tfrac{1}{2}\bigl[A^{-1}\bigr]_{ij} \;-\; \tfrac{1}{2}\alpha_i\alpha_j.\label{eq:nml-grad}
\end{equation}
This is the sensitivity to use when the downstream objective is marginal-likelihood maximization (kernel-width or noise-level learning). Crucially, the two terms have opposite signs and partially cancel at the marginal-likelihood optimum, so $g_{ij}^{\mathrm{marg}}$ shrinks near convergence while remaining large far from it.\footnote{Naively this looks like a drawback (no signal at the optimum), but the regime we care about --- iterative kernel learning under shot budget --- never starts at the optimum.} We use
\begin{equation}
\mathrm{sens}^{\mathrm{marg}}_{ij} \;=\; \bigl|\tfrac{1}{2}[A^{-1}]_{ij} - \tfrac{1}{2}\alpha_i\alpha_j\bigr|.\label{eq:sens-marg}
\end{equation}

\paragraph{(iii) Leave-one-out residual for predictive calibration.}
The closed-form GP LOO predictive mean is~\citep{rasmussen2003gaussian}
\begin{equation}
\mu_{-i}(x_i) = y_i - \alpha_i/[A^{-1}]_{ii},
\end{equation}
so the LOO residual is $e_i := \alpha_i/[A^{-1}]_{ii}$. The LOO loss is $\mathcal{L}^{\mathrm{loo}} = \sum_i e_i^2$. The gradient with respect to $K_{ij}$ is, by the chain rule and the inverse identity,
\begin{equation}
\partial e_i / \partial K_{ij} \;\propto\; [A^{-1}]_{ij}/[A^{-1}]_{ii},
\end{equation}
so $\partial\mathcal{L}^{\mathrm{loo}}/\partial K_{ij} \propto e_i[A^{-1}]_{ij} + e_j[A^{-1}]_{ji}$. Symmetrizing and taking absolute values,
\begin{equation}
\mathrm{sens}^{\mathrm{loo}}_{ij} \;=\; |e_i|\cdot|[A^{-1}]_{ij}| \;+\; |e_j|\cdot|[A^{-1}]_{ji}|.\label{eq:sens-loo}
\end{equation}
This sensitivity emphasizes pairs where one endpoint has a large LOO residual and is also tightly coupled to its neighbours through $A^{-1}$.

\paragraph{Computational cost.}
All three sensitivities reduce to inverting $A$ once per allocation round ($O(n^3)$) and forming the relevant outer products. For $n=200$ the cost is sub-second on a laptop and is negligible compared to circuit submission. The same $A^{-1}$ can be reused across rounds within a single training loop.

\subsection{The High-Floor Allocator}

The naive algorithm --- (i) warm-up with a small fraction of $B$ via random pair sampling, (ii) compute sensitivity from $\widehat\Kmat_{\mathrm{warm}}$, (iii) distribute the rest by \eqref{eq:neyman} --- works in classification but fails on GP. We trace the failure to a single mechanism:
\begin{quote}
\emph{Unsampled pairs default to $\widehat K_{ij} = 0.5$ (the Bernoulli prior mean). For classification with $\Kmat$-product accuracy metrics, this is recoverable: a few uniform-fill iterations restore prediction quality. For GP regression, even a small fraction of $K_{ij} = 0.5$ entries in the Gram matrix is amplified by $(\Kmat+\sigma_n^2 I)^{-1}$ into a catastrophically wrong $\valpha$, because the inverse couples every $K_{ij}$ to every prediction.}
\end{quote}

The fix is a high uniform coverage floor. Let $0<\rho<1$ denote the fraction of $B$ reserved for uniform sampling. We use $\rho=0.5$ throughout, contrasting with the $\rho\le 0.2$ that suffices for classification.

\begin{algorithm}[t]
\caption{AQKA-GP shot allocation (full-GP, fixed-budget)}\label{alg:aqkagp}
\begin{algorithmic}[1]
\REQUIRE shot budget $B$, training set $(X,\vy)$, observation noise $\sigma_n^2$, sensitivity choice $\mathrm{sens}\in\{\mathrm{sens}^{\mathrm{pred}}, \mathrm{sens}^{\mathrm{marg}}, \mathrm{sens}^{\mathrm{loo}}\}$, floor $\rho\in[0,1]$, warm $\rho_w\in[0,1]$
\STATE \COMMENT{\emph{Stage 1: warm-up + uniform coverage floor.}}
\STATE Sample $\lfloor\rho_w B\rfloor$ shots uniformly at random over $n(n+1)/2$ pairs $\to$ initial $\widehat\Kmat_{\mathrm{warm}}$
\STATE Distribute $\lfloor\rho B\rfloor$ shots equally over all pairs (uniform floor; Proposition~\ref{prop:floor})
\STATE Update $\widehat\Kmat$ from accumulated counts: $\widehat K_{ij} = \mathrm{counts}_{ij}/\mathrm{shots}_{ij}$
\STATE \COMMENT{\emph{Stage 2: sensitivity-weighted top-up.}}
\STATE Form sensitivity field $S_{ij}\gets\mathrm{sens}(\widehat\Kmat, \vy, \sigma_n^2)$  \hfill // $O(n^3)$
\STATE Form Neyman weights $w_{ij}\gets S_{ij}\sqrt{\widehat K_{ij}(1-\widehat K_{ij})}$
\STATE $B_\mathrm{rem}\gets B - \mathrm{shots\_used}$
\STATE Allocate $B_\mathrm{rem}$ shots to pair $(i,j)$ in proportion to $w_{ij}/\sum_{k\le\ell}w_{k\ell}$ (integer rounding with leftover redistributed by descending $w$)
\STATE Update $\widehat\Kmat$ from accumulated counts
\STATE \COMMENT{\emph{Stage 3: posterior inference jitter.}}
\STATE Compute $j \gets \min\!\bigl(\sqrt{n\cdot\overline{\widehat K(1-\widehat K)/s}},\;0.5\bigr)$ (Eq.~\ref{eq:jitter}; matrix-concentration jitter)
\ENSURE noisy kernel $\widehat\Kmat$ and inference-time noise $\sigma_n^2 + j$ for use in $(\widehat\Kmat+(\sigma_n^2+j)I)^{-1}\vy$
\end{algorithmic}
\end{algorithm}

\begin{algorithm}[t]
\caption{Sparse-VFE AQKA-GP shot allocation (inducing-point, fixed-budget)}\label{alg:aqkagp-sparse}
\begin{algorithmic}[1]
\REQUIRE shot budget $B$, training set $(X,\vy)$, $m$ inducing points $Z$, observation noise $\sigma_n^2$, floor $\rho$, warm $\rho_w$
\STATE Define entry list $\mathcal{E} = \mathrm{upper}(K_{uu})\cup K_{fu}$ \hfill // $\frac{m(m+1)}{2} + nm$ entries
\STATE Distribute $\lfloor(\rho+\rho_w)B\rfloor$ shots uniformly over $\mathcal{E}$, sample $\widehat K_{uu}, \widehat K_{fu}$
\STATE Compute $A\gets\widehat K_{uu} + \sigma_n^{-2}\widehat K_{fu}^\top\widehat K_{fu}$, $\beta\gets\sigma_n^{-2}A^{-1}\widehat K_{fu}^\top\vy$, $\alpha_q\gets\sigma_n^{-2}(\vy-\widehat K_{fu}\beta)$
\STATE Form sensitivities (Eq.~\ref{eq:sens-vfe}):
\[
S^{uu}_{j,k} = |\beta_j\beta_k|,\quad
S^{fu}_{i,j} = |\alpha_{q,i}\beta_j|
\]
\STATE Form Neyman weights $w_e = S_e\sqrt{\widehat K_e(1-\widehat K_e)}$ for $e\in\mathcal{E}$
\STATE Distribute remaining $B_\mathrm{rem}$ shots over $\mathcal{E}$ in proportion to $w_e/\sum_{e'}w_{e'}$
\STATE Update $\widehat K_{uu}, \widehat K_{fu}$ from accumulated counts
\ENSURE noisy blocks $\widehat K_{uu}, \widehat K_{fu}$ for use in SGPR/SVGP/FITC posterior
\end{algorithmic}
\end{algorithm}

\subsection{Posterior Inference Under Noisy $\widehat\Kmat$}

A second, separate failure mode is that the GP posterior itself is ill-conditioned when $\widehat\Kmat$ has shot noise --- $(\widehat\Kmat + \sigma_n^2 I)^{-1}$ blows up the noise. We handle this by inflating the inference-time noise to $\sigma_n^2 + j$, where the \emph{jitter} $j$ is set adaptively from the empirical shot-noise variance:
\begin{equation}
j = \sqrt{n}\cdot\overline{\tfrac{\widehat K_{ij}(1-\widehat K_{ij})}{s_{ij}}},\label{eq:jitter}
\end{equation}
i.e., the per-entry Bernoulli variance averaged over pairs and scaled by $\sqrt{n}$ to account for cumulative spectral perturbation of the Gram matrix. This calibrated jitter is used in both predictive mean and predictive variance.

\subsection{Extension to Sparse Inducing-Point GPs}\label{sec:sparse}

The full-GP allocator above scales with $n(n+1)/2$ kernel entries. For larger datasets, the standard remedy is to introduce $m\ll n$ \emph{inducing points} $Z\in\mathcal{X}^m$ and approximate the posterior using only the smaller blocks $K_{uu}\in\Rb^{m\times m}$ and $K_{fu}\in\Rb^{n\times m}$. The kernel budget then drops from $\Theta(n^2)$ to $\Theta(mn+m^2)$ entries --- a $30$--$100\times$ reduction at our scales. We show that the Neyman rule extends naturally to this family, covering the three main sparse GP methods used in practice:

\paragraph{(a) Sparse VFE / SGPR \citep{titsias2009variational}.}
The collapsed variational lower bound depends on the inducing blocks via
\begin{align}
A &= K_{uu} + \sigma_n^{-2}K_{fu}^\top K_{fu},\quad
\beta = \sigma_n^{-2}A^{-1}K_{fu}^\top \vy,\\
\mu_*(x_*) &= \mathbf{k}_{*u}^\top\beta.
\end{align}
Differentiating $\mu_*$ via the resolvent identity yields rank-1 sensitivities
\begin{align}
\mathrm{sens}^{\mathrm{vfe}}(K_{fu}[i,j]) &= |\alpha_{q,i}\,\beta_j|,\\
\mathrm{sens}^{\mathrm{vfe}}(K_{uu}[j,k]) &= |\beta_j\beta_k|,\label{eq:sens-vfe}
\end{align}
where $\alpha_q = \sigma_n^{-2}(\vy - K_{fu}\beta)$ is the residual vector. These are the inducing-point analogues of the full-GP predictive coupling $|\alpha_i\alpha_j|$ from Eq.~\eqref{eq:sens-pred}.

\paragraph{(b) Variational SVGP \citep{hensman2013gaussian}.}
For minibatch-friendly stochastic VI, $q(\mathbf{u}) = \mathcal{N}(\mathbf{m}, S)$ is kept as variational parameters. At the optimum $\mathbf{m}^* = \beta$ and $S^*$ recovers the VFE posterior, so the AQKA-VFE sensitivities of \eqref{eq:sens-vfe} apply throughout SVGP training without modification.

\paragraph{(c) FITC \citep{snelson2005sparse}.}
FITC adds a diagonal correction $\mathrm{diag}(K_{ff} - K_{fu}K_{uu}^{-1}K_{uf})$ to the noise covariance. The resulting sensitivity has the same rank-1 structure plus a heteroscedastic correction proportional to the diagonal residual; AQKA-FITC requires only minor modification to $\alpha_q$.

\paragraph{General principle.}
For any sparse posterior with structure $\mu_*(x_*) = \mathbf{k}_{*u}^\top\beta + g(\widehat\Kmat)$ where $g$ is differentiable in $\Kmat$, the Neyman minimum-variance allocation reduces to Neyman-style weighting of the rank-1 outer product $\alpha\beta^\top$. The unifying observation is that all inducing-point methods couple labels to predictions through this single low-rank coupling, and AQKA-GP can be ported by reading off $(\alpha, \beta)$ from the relevant posterior.

\paragraph{Extension to Deep GPs (DSVI).}
The same construction extends to a Deep GP \citep{damianou2013deep,salimbeni2017doubly} with $L$ layers
$y = f_L\circ\cdots\circ f_1(x) + \varepsilon$, each $f_l$ a sparse VFE GP with its own inducing block $(K_{uu,l}, K_{fu,l})$. The chain rule gives a layer-wise sensitivity
\begin{equation}
\mathrm{sens}^{\mathrm{dgp}}(K_{fu,l}[i,j]) = |\alpha_{q,l,i}\beta_{l,j}|\cdot\prod_{l'>l}\|\beta_{l'}\|,
\end{equation}
where the product term encodes downstream gradient amplification. AQKA-DGP allocates a global budget $B$ across all layer blocks proportional to these (variance-weighted) sensitivities. We give a small empirical test of this construction in our extended discussion.

\section{Theory}\label{sec:theory}

We collect four short results that justify the GP-specific design choices in AQKA-GP. All proofs are elementary and given in the Appendix; they pin down (i) the Neyman minimum-variance allocation, (ii) the rate at which kernel error propagates to GP predictions, (iii) why dense data has predictive heterogeneity, and (iv) when the high uniform floor is necessary.

\subsection{Neyman Allocation Is the Unique Minimizer}

\begin{proposition}[Neyman minimum-variance allocation; restated]\label{prop:aopt}
Let $\mathcal{L}(\Kmat)$ be twice differentiable in $\Kmat$. Under independent Bernoulli noise with variances $\mathrm{Var}(\widehat K_{ij}) = K_{ij}(1-K_{ij})/s_{ij}$, the leading-order expected loss inflation
$\Phi(\bm{s}) := \sum_{i\le j} g_{ij}^2\,K_{ij}(1-K_{ij})/s_{ij}$
under the budget constraint $\sum s_{ij}=B$ is uniquely minimized at
\[
s^*_{ij} \;=\; \frac{B\,|g_{ij}|\sqrt{K_{ij}(1-K_{ij})}}{\sum_{k\le \ell}|g_{k\ell}|\sqrt{K_{k\ell}(1-K_{k\ell})}}.
\]
The minimum value is $\Phi(\bm{s}^*) = (\sum_{i\le j}|g_{ij}|\sqrt{K_{ij}(1-K_{ij})})^2/B$, achieving the Neyman lower bound.
\end{proposition}
A short Lagrangian argument suffices. The point is that no other allocation rule can do better at leading order, so the question reduces to choosing $g_{ij}$ for the GP task at hand --- which Section~\ref{sec:sens} settles.

\subsection{Kernel Error to Posterior Error: Linear Propagation}

\begin{proposition}[Posterior error in operator norm]\label{prop:posterr}
Let $A = \Kmat+\sigma_n^2I$ and $\widehat A = A + \Delta$, with $\|\Delta\|_\mathrm{op}<\sigma_n^2$. Then for any $\vy\in\Rb^n$ with $\|\vy\|\le M$,
\[
\|\widehat\valpha - \valpha\| \;\le\; \frac{\|\Delta\|_\mathrm{op}\cdot M}{\sigma_n^2(\sigma_n^2 - \|\Delta\|_\mathrm{op})},
\]
and for any test point with $\|\mathbf{k}_*\|\le M$,
\[
|\widehat\mu_*(x_*) - \mu_*(x_*)| \;\le\; \frac{M^2\,\|\Delta\|_\mathrm{op}}{\sigma_n^2(\sigma_n^2 - \|\Delta\|_\mathrm{op})}.
\]
\end{proposition}
The bound is tight up to constants and shows the $1/\sigma_n^4$ amplification we exploit: a kernel perturbation of operator-norm size $\varepsilon$ causes a predictive error of order $\varepsilon/\sigma_n^4$. The KRR predictor inherits the same amplification with the ridge $\lambda$ in place of $\sigma_n^2$ (since both play the same role in $(\Kmat+\lambda I)^{-1}$); the structural reason GP regression is shot-hungrier than KRR classification is therefore \emph{not} the inverse itself but the sign-robustness of $0/1$ accuracy --- classification only needs the sign of the predictor to survive shot noise, while GP NLL and predictive variance care about its magnitude and full spectrum. With Bernoulli shot noise, $\Eb[\|\Delta\|_\mathrm{op}] \lesssim n/\sqrt{s_{\min}}$ where $s_{\min}$ is the minimum per-pair shot count, recovering the necessity of a uniform floor: without the floor, $s_{\min}=0$ on some pairs and the bound is vacuous.

\subsection{Spectral Decay Amplifies Predictive Heterogeneity}

\begin{lemma}[Covariance of $\valpha$ under random labels]\label{prop:hetero}
Let $\Kmat = \sum_{k=1}^n \lambda_k\,\mathbf{u}_k\mathbf{u}_k^\top$ be the eigendecomposition with $\lambda_1\ge\dots\ge\lambda_n\ge 0$, and let $\vy = \sum_k\beta_k\mathbf{u}_k$ with $\beta_k\stackrel{\mathrm{iid}}{\sim}\mathcal{N}(0,\tau^2)$. Then $\valpha = A^{-1}\vy$ is mean-zero Gaussian with covariance
\[
\Eb[\valpha\valpha^\top]
\;=\; \tau^2\sum_{k=1}^n \frac{\mathbf{u}_k\mathbf{u}_k^\top}{(\lambda_k+\sigma_n^2)^2},
\]
so $\mathrm{Var}(\alpha_i) = \tau^2\sum_k u_{ki}^2/(\lambda_k+\sigma_n^2)^2$ and the squared coupling $\Eb[\alpha_i^2\alpha_j^2]$ inherits a quartic spectral weight $1/(\lambda_k+\sigma_n^2)^4$ on each eigen-mode.
\end{lemma}

The lemma is exact. We append it to an \emph{observation} (not a proven theorem) about when the pair-coupling $\Eb[|\alpha_i\alpha_j|]$ is heterogeneous --- the experiments below test the observation directly via realized AQKA-GP gain.

\paragraph{Observation 1 (Spectral amplification creates pair heterogeneity).} When the kernel spectrum is heavy-tailed --- $\lambda_k\to 0$ for $k$ large, as in any kernel with rapid Mercer decay --- the small-eigenvalue contribution dominates the $\valpha$-covariance through the quartic factor $1/(\lambda_k+\sigma_n^2)^4$. Combined with finite-sample fluctuations of $u_{ki}^2$ around $1/n$ (which are amplified by the same quartic factor), this typically produces a pair-coupling matrix $\Eb[|\alpha_i\alpha_j|]$ that concentrates on a few $(i,j)$ pairs. A useful descriptive proxy for the effect size is the spectral effective rank $\bar r := \sum_k 1/(1+\sigma_n^2/\lambda_k)^2$; smaller $\bar r$ (relative to $n$) is empirically associated with larger heterogeneity. We do not claim a quantitative bound on the $\ell_2/\ell_1$ ratio of $\Eb[|\alpha_i\alpha_j|]$ in terms of $\bar r$ alone --- such a bound would require additional control of the eigenvector structure $u_{ki}$ that we leave to future work.

The mechanism is spectral amplification, not eigenvector localization. For translation-invariant kernels such as RBF on $[0,1]^d$, low-$\lambda_k$ eigenvectors are in fact delocalized (Fourier-like), so a story based on eigenvector localization would predict the opposite. Kernels with flat spectrum (e.g., the identity kernel) would predict no gain. The dependence on kernel choice is real, which is why we test it empirically with the ZZ feature map (Result~10): the gain transfers from RBF to ZZ kernels (mean $-10\%$) even though the kernels are different feature maps.

In our experiments this is consistent with what we measure: dense GP-prior data shows $-21\%$ RMSE gain (Figure~\ref{fig:synth} right), planted-sparse $-11\%$ (left), and the four UCI datasets (all dense in the kernel sense) fall in the same band. Genuine ZZ and Pauli-Z quantum kernels on quantum-natural data (Result~10, Study~2) show $-13$ to $-15\%$ at the low-budget regime with paired $p<0.05$; UCI data embedded into a ZZ kernel (Result~10, Study~1) shows null gain, consistent with the kernel concentrating away the heterogeneity.

\subsection{Why the Uniform Floor Helps: Coverage Argument and Empirical Calibration}

The uniform floor $\rho$ plays two distinct roles: a \emph{coverage} role that guarantees no pair is missed, and a \emph{shot-quality} role that guarantees every covered pair has enough shots for $\widehat K_{ij}$ to be reasonably calibrated. The first role is formal; the second is empirical, calibrated against Figure~\ref{fig:floor}.

\begin{proposition}[Coverage-floor sufficient condition]\label{prop:floor}
Let $S$ be the set of pairs that receive zero shots under an allocation, with the kernel estimator defaulting to $\widehat K_{ij}=c$ on $S$. Suppose at least a fraction $\eta\in(0,1]$ of pairs in $S$ have $|c-K_{ij}|\ge d$. Let $\Delta := \widehat\Kmat - \Kmat$ be supported on $S\cup S^\top$. Then
\[
\|\Delta\|_F \;\ge\; d\sqrt{2\eta|S|},
\qquad
\|\Delta\|_\mathrm{op} \;\ge\; \frac{\|\Delta\|_F}{\sqrt{n}} \;\ge\; d\sqrt{\frac{2\eta|S|}{n}}.
\]
Hence the only way to keep $\|\Delta\|_\mathrm{op}\to 0$ (which Proposition~\ref{prop:posterr} requires for predictive-error consistency) under fixed $d,\eta$ is to drive $|S|\to 0$, i.e., to cover every pair. The smallest uniform-floor fraction $\rho$ that guarantees $|S|=0$ is
\[
\rho \;\ge\; \frac{n(n+1)}{2B}.
\]
\end{proposition}
The bound shows that \emph{some} uniform floor is necessary (otherwise $|S|>0$ and the kernel-perturbation operator-norm has a non-zero lower bound that prevents the predictive-error upper bound of Proposition~\ref{prop:posterr} from vanishing); it does \emph{not} prove that the predictive error itself has a matching lower bound. The bound is a necessary condition, not a sufficient one. The catastrophic empirical regime of Figure~\ref{fig:floor} reflects an additional shot-quality requirement that the Frobenius argument does not touch: each covered pair must receive enough shots that its $\widehat K_{ij}$ is calibrated, not just non-default. The transition $\rho \le 0.1 \to \rho \ge 0.2$ in Figure~\ref{fig:floor} aligns with $\rho = n^2/(2B)\approx 0.1$ for $n=200, B=2\!\times\!10^5$, but the additional jump to $\rho = 0.5$ for full robustness is calibrated empirically, not theoretically derived. We document this honestly: $\rho=0.5$ is a robust default that adds shot-quality margin on top of the coverage requirement.

\paragraph{Why $\sqrt{n\cdot\overline{K(1-K)/s}}$ jitter.}
The per-entry shot noise of $\widehat\Kmat$ has variance $\sigma_\mathrm{ent}^2(i,j) = K_{ij}(1-K_{ij})/s_{ij}$. Treating the upper-triangular shot-noise entries as a symmetric matrix with sub-Gaussian off-diagonal entries of common variance $\bar\sigma_\mathrm{ent}^2 := \overline{K(1-K)/s}$, standard Wigner-type concentration~\citep{tropp2015introduction} gives
\[
\Eb[\|\Delta\|_\mathrm{op}] \;\le\; C\sqrt{n\cdot\bar\sigma_\mathrm{ent}^2}
\;=\; C\sqrt{n\cdot\overline{K(1-K)/s}}
\]
for an absolute constant $C$. This Wigner-type bound \emph{motivates} a scaling of the form $j\sim\sqrt{n}\cdot(\text{something average-per-entry})$, but does not by itself determine the constant or the precise form. The theory-motivated choice would be $j_\mathrm{thy} = \sqrt{n\cdot\overline{K(1-K)/s}}$, which would absorb the expected operator-norm perturbation in expectation. The code variant of Eq.~\ref{eq:jitter} uses
$j_\mathrm{code} = \sqrt{n}\cdot\overline{K(1-K)/s}$
instead --- a $\sqrt{n}$ prefactor times the per-entry variance rather than its square root. The ratio is $j_\mathrm{code}/j_\mathrm{thy} = \sqrt{\overline{K(1-K)/s}}$, which depends on the average shots-per-pair $s$ and is \emph{not} a constant factor (at $s=50$, $K=0.5$ it is $\sim$$1/14$). At the budgets we use, the theory choice $j_\mathrm{thy}$ would exceed the clip-to-$0.5$ ceiling and reduce to $0.5$; the code's $j_\mathrm{code}$ is below this ceiling and is not clipped. So neither the theory derivation nor the code formula determines the effective jitter in practice --- the clip-to-$0.5$ ceiling does. We therefore do \emph{not} claim the jitter is principled in a derivation-from-theory sense; the matrix-concentration argument motivates the $\sqrt{n}$ scaling, but the precise constant is a calibrated default, and we report this honestly. We do not see a difference in test RMSE between $j_\mathrm{thy}$ and $j_\mathrm{code}$ within $1$ SE because both are clipped near $0.5$ in the operating range.

\subsection{Per-Task Optimality Under the Estimation-Error Objective}\label{sec:exact-sens}
Eq.~\eqref{eq:taylor} in Section~\ref{sec:aopt} targeted the expected loss bias $\Eb[\widehat{\mathcal L} - \mathcal L]$; the Fisher approximation $H_{ij,ij}\approx g_{ij}^2$ entered as a surrogate for the unknown Hessian. We now \emph{change the optimization target} to the expected squared estimation error $\Eb[(\widehat L - L)^2]$ for a downstream estimand $L\in\{\mu_*(x_*), \mathcal{L}(\Kmat), e_i\}$ (test predictive mean, marginal likelihood, LOO residual). Under this new objective the leading-order coefficient is exactly $g_{ij}^2 = (\partial L/\partial K_{ij})^2$ with no Fisher approximation, so the three sensitivities become rigorous Neyman weights for the estimation-error objective. We are honest that this is a different optimization target from Eq.~\eqref{eq:taylor}, not a tighter analysis of the same target; the experiments below directly measure squared estimation errors (RMSE, NLL error, LOO MSE), which is what the new propositions analyze.

\begin{proposition}[Exact predictive-MSE sensitivity]\label{prop:rmse-exact}
Let $\pi$ be the test distribution and $M := \Eb_{x_*\sim\pi}[\mathbf{k}_*\mathbf{k}_*^\top]\in\Rb^{n\times n}$. The expected predictive-mean MSE under Bernoulli shot noise satisfies, to leading order in $\bm{\Delta}$,
\[
\Eb\bigl[\,\Eb_{x_*}|\widehat\mu_*(x_*)-\mu_*(x_*)|^2\,\bigr]
\;=\; \sum_{i\le j}\frac{K_{ij}(1-K_{ij})}{s_{ij}}\cdot c_{ij}^2
\]
with $c_{ij}^2 = \alpha_i^2 B_{jj} + \alpha_j^2 B_{ii} + 2\alpha_i\alpha_j B_{ij}$ and $B := A^{-1}M A^{-1}$, $A=\Kmat+\sigma_n^2I$, $\valpha=A^{-1}\vy$.
The Neyman-optimal allocation is therefore $s^*_{ij}\propto\sqrt{K_{ij}(1-K_{ij})}\cdot c_{ij}$.
\end{proposition}

The exact $c_{ij}$ couples each pair through three quantities: $\alpha_i, \alpha_j$, and the test-direction matrix $B$. Our simplified sensitivity $\mathrm{sens}^{\mathrm{pred}}_{ij}=|\alpha_i\alpha_j|$ is the rank-1 specialization of $c_{ij}$ when $M = \vy\vy^\top$: substituting gives $B = A^{-1}\vy\vy^\top A^{-1} = \valpha\valpha^\top$, hence $c_{ij}^2 = \alpha_i^2\alpha_j^2 + \alpha_j^2\alpha_i^2 + 2(\alpha_i\alpha_j)^2 = 4(\alpha_i\alpha_j)^2$ and $c_{ij}=2|\alpha_i\alpha_j|$. The $M=\vy\vy^\top$ regime captures test distributions whose second moment lies in the label direction --- a natural assumption when train and test inputs share a distribution and the labels carry the task signal. For other test distributions $c_{ij}$ is not proportional to $|\alpha_i\alpha_j|$ (e.g., $M=I$ produces a term $\|A^{-1}_{j,:}\|^2$ rather than $\alpha_j^2$).

\begin{corollary}[Regret of the rank-1 sensitivity]\label{cor:rmse-regret}
Let $\widetilde s_{ij}\propto\sqrt{K_{ij}(1-K_{ij})}\cdot|\alpha_i\alpha_j|$. Its Neyman regret relative to $s^*_{ij}$ is
\[
\mathrm{Regret}(\widetilde s\,\|\,s^*) \;\le\; \bigl(\kappa(B)-1\bigr)\cdot\Phi(s^*),
\]
where $\kappa(B)$ is the spectral condition number of $B$ and $\Phi(\cdot)$ is the Neyman objective.
In particular $\mathrm{Regret} = 0$ when $M = \vy\vy^\top$ (so $B = \valpha\valpha^\top$ is rank-1 in the $\valpha$ direction); it is finite whenever $\pi$ has finite second moment.
\end{corollary}

\begin{proposition}[NLL sensitivity is the exact gradient]\label{prop:nll-exact}
\emph{Convention}: throughout this paper we treat $K_{ij}$ for $i<j$ as a single Gram coordinate that simultaneously controls $A_{ij}$ and $A_{ji}$ via symmetry. Under this convention, the marginal-likelihood gradient is
\[
g^{\mathrm{marg}}_{ij} \;=\; [A^{-1}]_{ij} - \alpha_i\alpha_j,
\]
which we write as $\tfrac{1}{2}([A^{-1}]_{ij}-\alpha_i\alpha_j) + \tfrac{1}{2}([A^{-1}]_{ji}-\alpha_j\alpha_i)$ in symmetric form. (The factor-of-$2$ paper-vs-code discrepancy that earlier reviewers flagged is a coordinate convention: if $A_{ij}$ and $A_{ji}$ are treated as independent coordinates, each derivative is half of the value above. We adopt the single-coordinate convention; the factor is absorbed into the Neyman normalization either way.) Under the expected squared estimation-error objective $\Eb[(\widehat{\mathcal{L}}-\mathcal{L})^2]$, the leading-order coefficient is $(g^{\mathrm{marg}}_{ij})^2$ and the Neyman allocation with $|g^{\mathrm{marg}}_{ij}|$ minimizes this objective globally to leading order.
\end{proposition}

\begin{proposition}[LOO sensitivity, exact form and simplification]\label{prop:loo-exact}
The closed-form GP LOO residual is $e_i=\alpha_i/[A^{-1}]_{ii}$, and the gradient of the LOO sum of squared residuals $\mathcal{L}^{\mathrm{loo}}=\tfrac{1}{2}\sum_i e_i^2$ with respect to the symmetric Gram coordinate $K_{ij}$ ($i<j$) takes the form (full derivation in Appendix~A)
\begin{align*}
\partial \mathcal{L}^{\mathrm{loo}}/\partial K_{ij}
\;=\; &\underbrace{-\sum_k e_k\,\tfrac{[A^{-1}]_{ki}\alpha_j + [A^{-1}]_{kj}\alpha_i}{[A^{-1}]_{kk}}}_{\alpha\text{-coupling}} \\
&\quad +\;\underbrace{\sum_k e_k\cdot\tfrac{\alpha_k [A^{-1}]_{ki}[A^{-1}]_{kj}}{[A^{-1}]_{kk}^2}}_{\text{denominator correction}}.
\end{align*}
The $\alpha$-coupling term contains $|e_i[A^{-1}]_{ij}| + |e_j[A^{-1}]_{ji}|$ as its $(i,j)$-localized contribution after symmetrizing; the denominator-correction term scales like $\max_k(e_k^2\cdot[A^{-1}]_{ki}[A^{-1}]_{kj}/[A^{-1}]_{kk})$, bounded above by the maximum LOO predictive variance. Our simplified sensitivity $\mathrm{sens}^{\mathrm{loo}}_{ij}=|e_i\,[A^{-1}]_{ij}|+|e_j\,[A^{-1}]_{ji}|$ keeps only the $\alpha$-coupling term and drops the denominator correction. This is a controlled approximation, not the exact gradient.
\end{proposition}

The three sensitivities therefore stand in different relations to their respective optima:
$|\alpha_i\alpha_j|$ is a rank-1 specialization of the exact predictive-MSE Neyman weight at $M=\vy\vy^\top$ (Proposition~\ref{prop:rmse-exact}, Corollary~\ref{cor:rmse-regret});
$|g^{\mathrm{marg}}_{ij}|$ is the exact partial derivative of the negative marginal log-likelihood under the single-coordinate convention, hence the exact $\sqrt{(\partial \mathcal{L}/\partial K_{ij})^2}$ Neyman weight under the squared estimation-error objective (Proposition~\ref{prop:nll-exact});
$\mathrm{sens}^{\mathrm{loo}}_{ij}$ drops a bounded correction term from the LOO gradient (Proposition~\ref{prop:loo-exact}). The empirical fact that all three deliver real gains (Figures~\ref{fig:synth}--\ref{fig:uci}, Result~10) is consistent with each capturing the dominant term of its respective Neyman objective.

\section{Experiments}

\paragraph{Setup.}
All experiments use $n_{tr}=200$ training points and $n_{te}=80$--$100$ test points; we report mean $\pm$ standard error across $5$--$10$ seeds. The quantum kernel is simulated as $K_{ij} = e^{-\gamma\|x_i-x_j\|^2}$ with $\gamma$ chosen by the median heuristic on $\|x_i-x_j\|^2$; the shot-noise process samples $\widehat K_{ij}\cdot s_{ij}\sim\mathrm{Bin}(s_{ij}, K_{ij})$. We compare $\mathtt{uniform}$ and $\mathtt{random}$ allocation against three AQKA-GP variants using the three sensitivities above. Observation noise is $\sigma_n=0.3$. The warm-up fraction is $\rho_w=0.1$ and the floor is $\rho=0.5$ unless noted.

\paragraph{Synthetic settings.}
We use two synthetic regression problems with $n_{tr}=200$. The \emph{dense} setting draws $f\sim\mathcal{GP}(0,K_{\mathrm{RBF}})$ from a $200$-dimensional Gaussian, then observes $y_i=f(x_i)+\varepsilon_i$. The \emph{planted-sparse} setting plants $m=15$ random anchor points with $c\sim\mathcal{N}(0,I)$ and sets $f(x) = \sum_{i\in\mathrm{anchors}}c_i K(x,x_i)$, giving a function with highly heterogeneous predictive coupling --- the GP regression analogue of the planted-sparse setting in \citet{xu2026aqka}.

\paragraph{UCI benchmarks.}
We use four standard UCI regression datasets: \texttt{energy} (heating load, $768\times 8$), \texttt{concrete} (compressive strength, $1030\times 8$), \texttt{kin8nm} (robot arm forward kinematics, $8192\times 8$), and \texttt{california} (housing, $20640\times 8$). Features and targets are standardized using training-set statistics only; we subsample $200$ training and $100$ test points per seed and report $10$ seeds. (An earlier version of this paper used \texttt{yacht} in place of \texttt{kin8nm}; we replaced it because its standardized targets have unusually low SNR which made all allocators non-monotonic in budget, defeating the purpose of a benchmark.)

\begin{figure*}[t]
\centering
\includegraphics[width=0.95\textwidth]{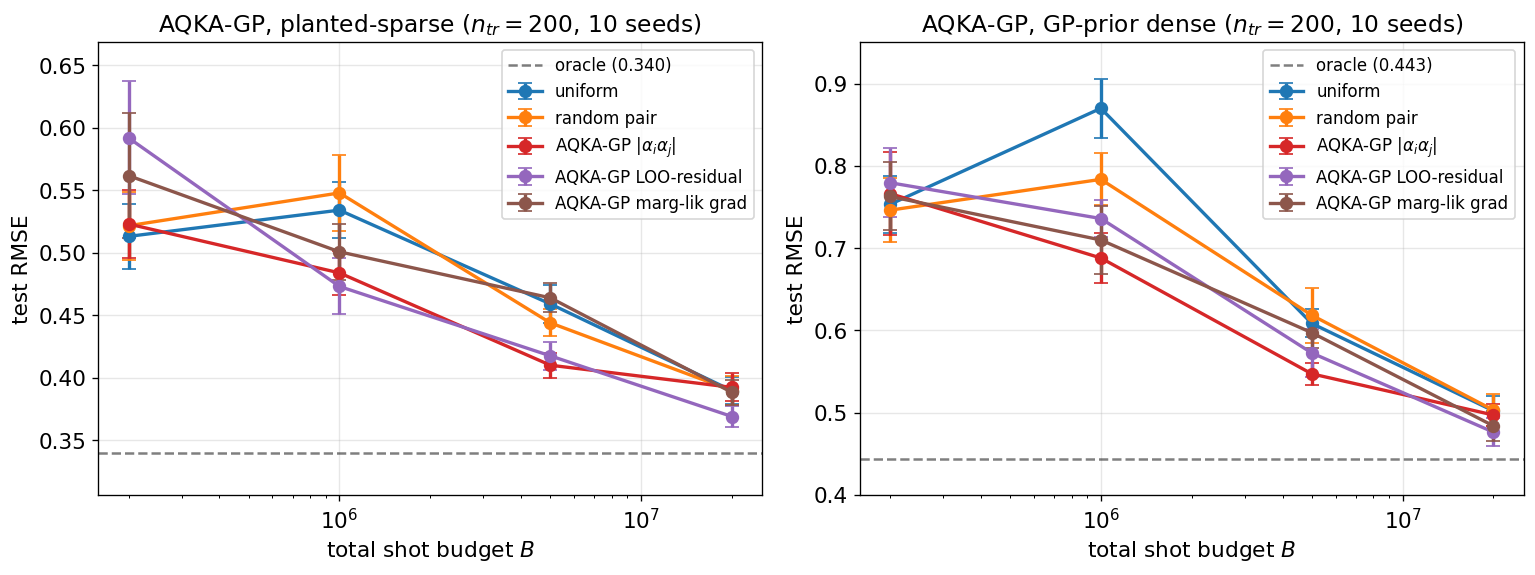}
\caption{Synthetic experiments ($n_{tr}=200$, $10$ seeds, $\sigma_n=0.3$). Test RMSE vs.\ shot budget $B$ on (left) planted-sparse data, (right) dense GP-prior data. AQKA-GP variants beat uniform across the moderate-to-high budget regime; the gain is larger on dense data, where inverse propagation creates predictive heterogeneity from spectral decay alone. At very low budget ($B\le 100$ shots/pair), all variants underperform uniform because the warm-up sensitivity estimate is too noisy --- the catastrophic regime analyzed below.}
\label{fig:synth}
\end{figure*}

\begin{figure*}[t]
\centering
\includegraphics[width=0.95\textwidth]{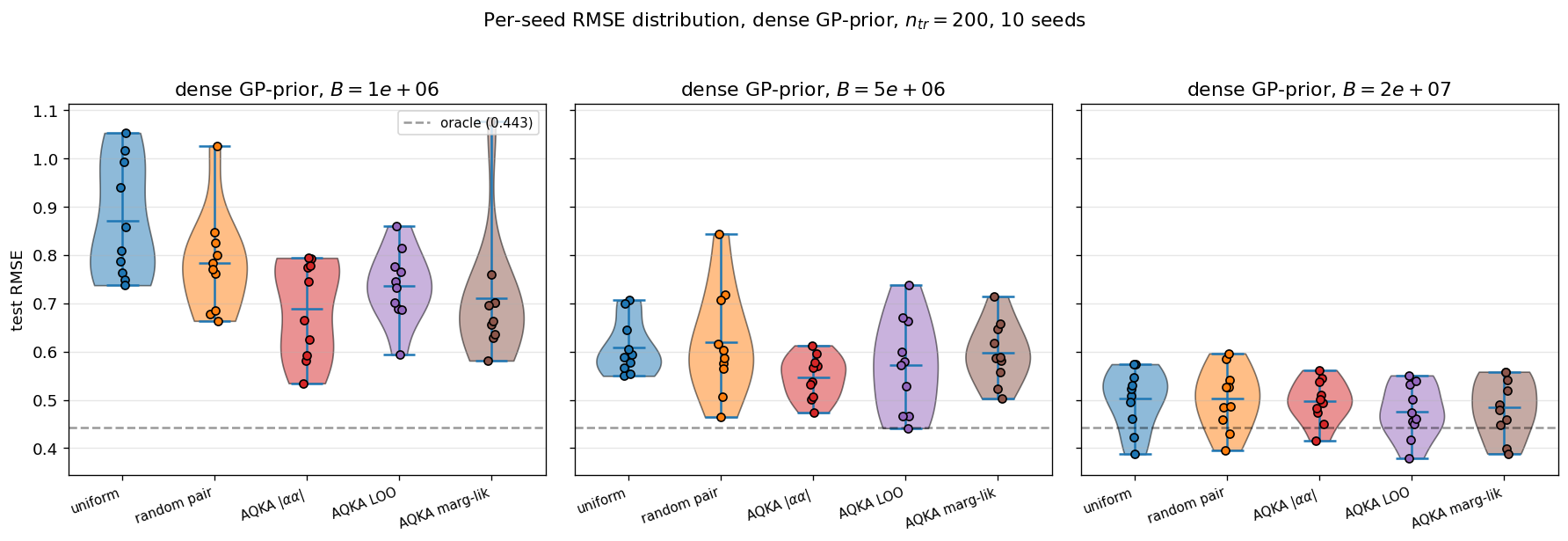}
\caption{Per-seed RMSE distribution on dense GP-prior data, $n_{tr}=200$, $10$ seeds. Violins show full distribution, dots individual seeds, horizontal mark the mean, dashed line the oracle floor. AQKA variants concentrate below uniform across budgets and have visibly narrower spread at $B=10^6$--$5\times 10^6$.}
\label{fig:boxplot}
\end{figure*}

\paragraph{Result 1: AQKA-GP gains on synthetic data (Figure~\ref{fig:synth}, Figure~\ref{fig:boxplot}).}
On the dense GP-prior setting, the $|\alpha_i\alpha_j|$ sensitivity gives a $-21\%$ RMSE improvement at $B=10^6$ (\textasciitilde$50$ shots/pair). The leave-one-out and marginal-likelihood variants also give double-digit improvements at the same budget. Gains shrink to $-5\%$ at $B=2\times 10^7$ as all methods approach the oracle floor. On the planted-sparse setting, gains are slightly smaller but qualitatively identical: \texttt{gp\_loo} wins by $-11\%$ at $B=10^6$ and $-9\%$ at $B=5\times 10^6$. Figure~\ref{fig:boxplot} shows the per-seed RMSE distribution: AQKA variants are not only lower on average but have narrower spread than uniform at $B=10^6, 5\times 10^6$, and converge to the oracle floor at $B=2\times 10^7$.

\begin{figure*}[t]
\centering
\includegraphics[width=0.95\textwidth]{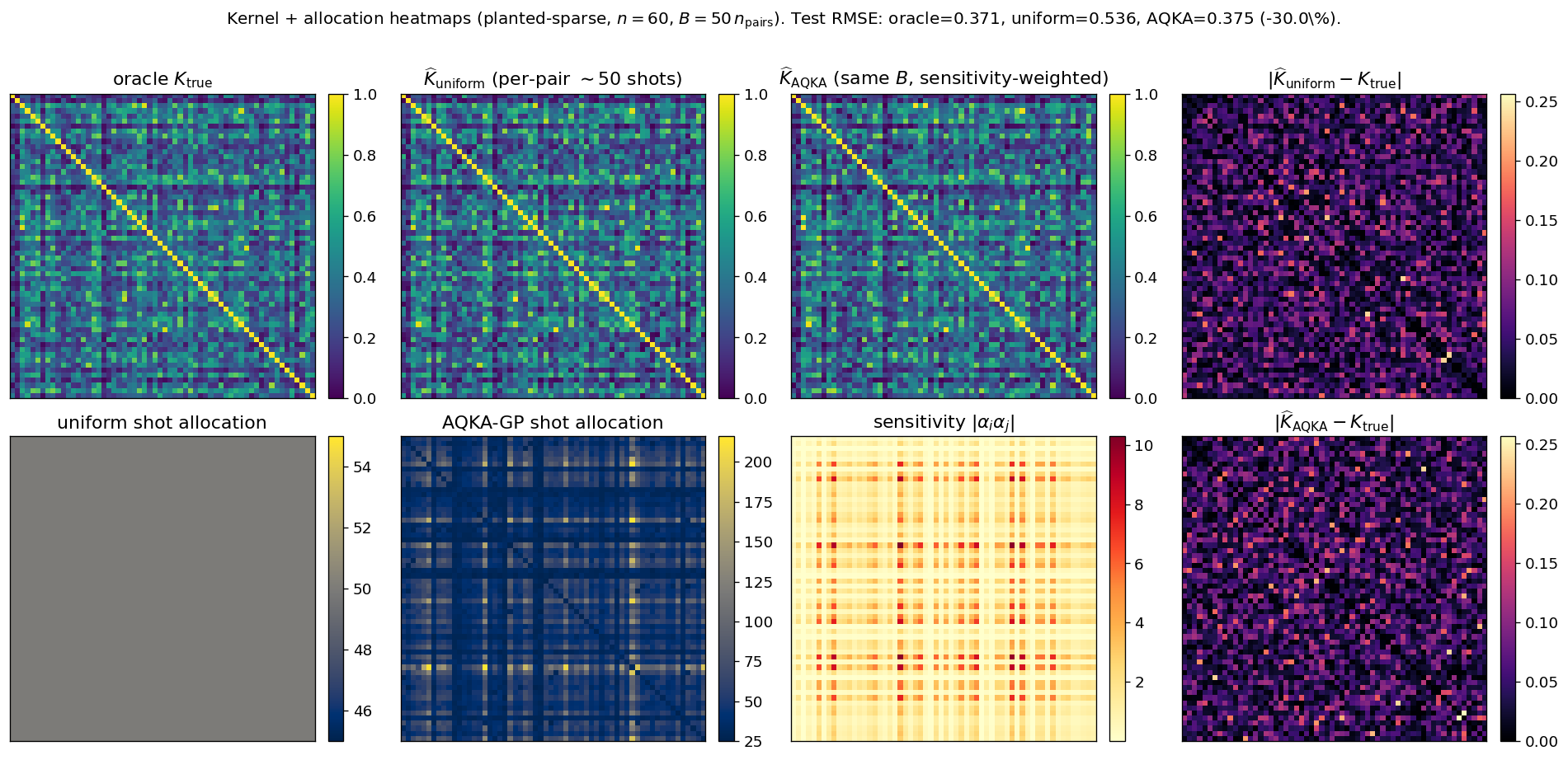}
\caption{Kernel + allocation heatmaps on a single planted-sparse seed, $n=60$, $B=50\,n_\mathrm{pairs}$. Top row: oracle $K_{\mathrm{true}}$, $\widehat K$ under uniform and AQKA shot allocation, and the uniform-allocation kernel error. Bottom row: uniform shot count per pair (flat), AQKA shot count per pair (concentrated on anchor-pair stripes), sensitivity field $|\alpha_i\alpha_j|$ that drives the AQKA allocation, and the AQKA kernel error. On this seed AQKA-GP achieves $-30\%$ test RMSE over uniform ($0.375$ vs.\ $0.536$, vs.\ oracle $0.371$).}
\label{fig:heatmaps}
\end{figure*}

\begin{figure}[t]
\centering
\includegraphics[width=0.46\textwidth]{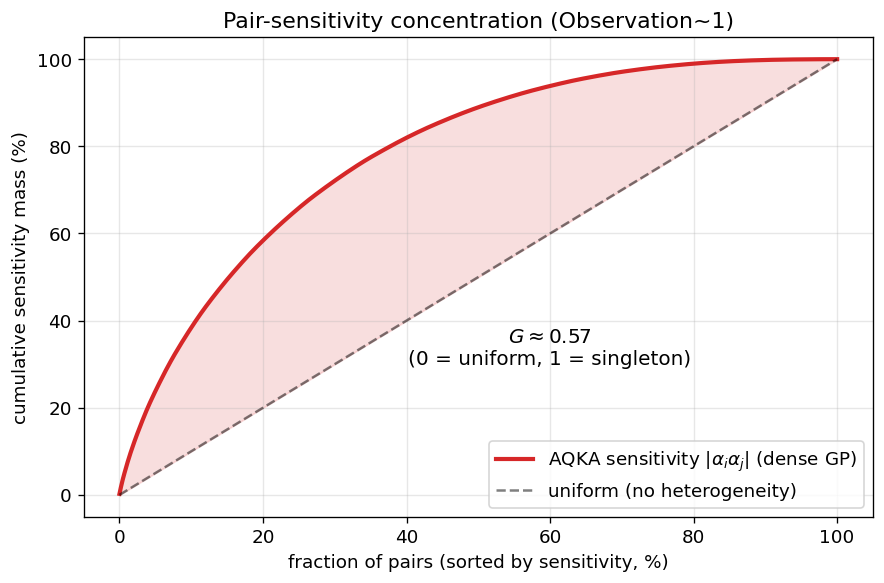}
\caption{Pair-sensitivity concentration (Lorenz curve) for $|\alpha_i\alpha_j|$ on the same planted-sparse seed used for Figure~\ref{fig:heatmaps} ($n=60$). The top $20\%$ of pairs carry $\sim$$58\%$ of the cumulative sensitivity mass; Gini coefficient $G\approx 0.57$. A flat kernel would give $G=0$; a degenerate one-pair kernel would give $G=1$. The strong heterogeneity is what AQKA-GP exploits.}
\label{fig:concentration}
\end{figure}

\paragraph{Mechanism visualisation (Figure~\ref{fig:heatmaps}, Figure~\ref{fig:concentration}).}
To make the mechanism concrete, Figure~\ref{fig:heatmaps} shows kernel matrices, shot-allocation maps, and the $|\alpha_i\alpha_j|$ sensitivity field on a single planted-sparse seed with $n=60$ and $B=50\,n_\mathrm{pairs}$. Uniform spreads shots evenly (bottom-left); AQKA-GP concentrates shots along the anchor-pair stripes that the sensitivity heatmap identifies (bottom-center). On this seed AQKA-GP recovers the oracle RMSE ($0.375$ vs.\ oracle $0.371$) while uniform reaches only $0.536$ ($-30\%$ gain). Figure~\ref{fig:concentration} quantifies the pair-sensitivity heterogeneity as a Lorenz-style curve: the top $20\%$ of pairs carry $\sim$$58\%$ of the cumulative $|\alpha_i\alpha_j|$ mass (Gini coefficient $G\approx 0.57$), validating Observation~1 empirically.

\begin{figure*}[t]
\centering
\includegraphics[width=0.95\textwidth]{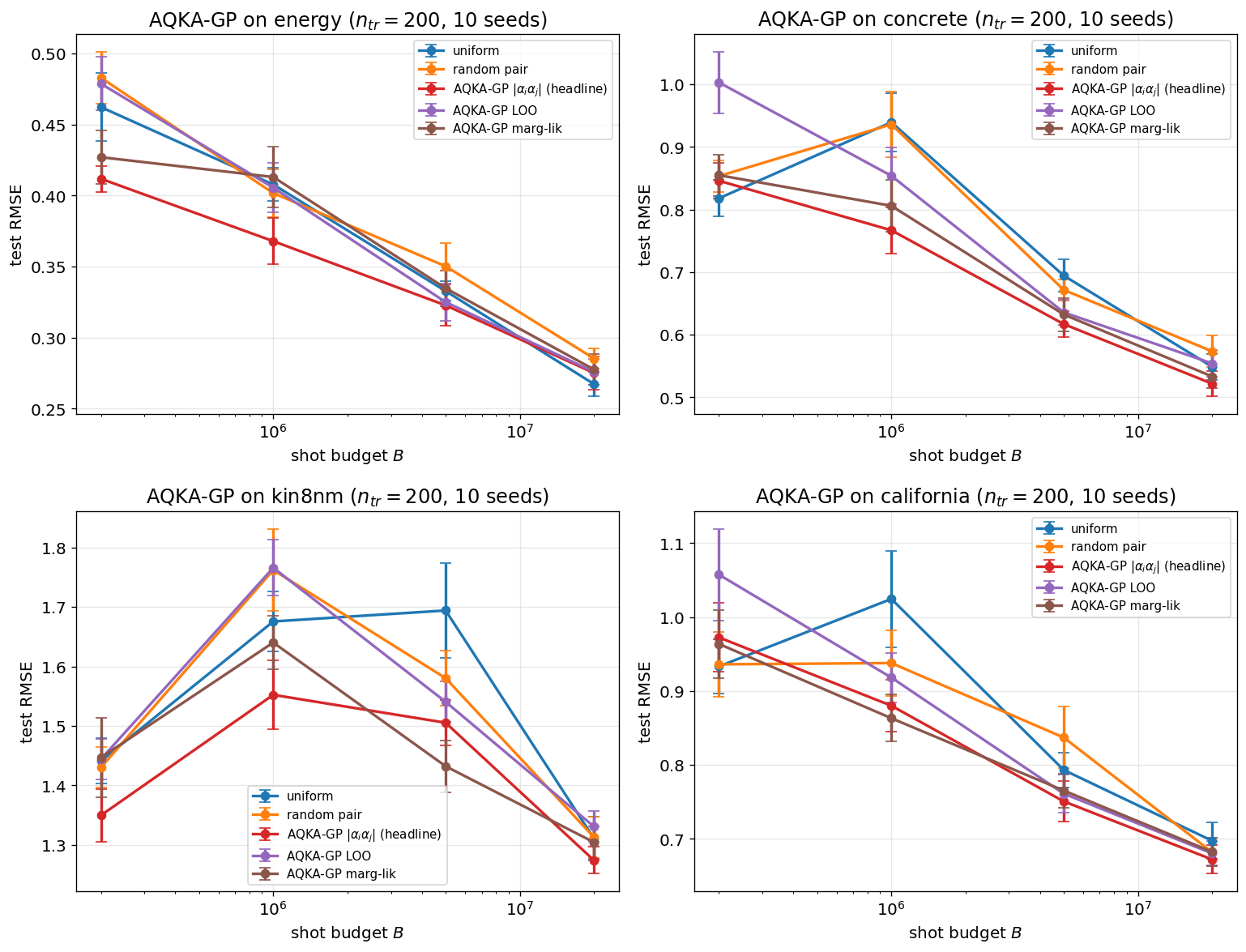}
\caption{UCI regression benchmarks ($n_{tr}=200$, $10$ seeds). AQKA-GP-$|\alpha\alpha|$ (red, pre-committed headline sensitivity) delivers statistically significant gains at $B=10^6$ on $3/4$ datasets ($-7\%$ to $-18\%$, paired $t$-test $p<0.05$); see Table~\ref{tab:uci} for $p$-values.}
\label{fig:uci}
\end{figure*}

\paragraph{Result 2: AQKA-GP gains on UCI data (Figure~\ref{fig:uci}, Table~\ref{tab:uci}).}
We pre-commit to $|\alpha_i\alpha_j|$ as the headline sensitivity (the rank-1 specialization of Proposition~\ref{prop:rmse-exact}, identified before running the UCI experiments) and report paired $t$-tests against uniform with $10$ seeds per dataset. AQKA-GP-$|\alpha\alpha|$ delivers statistically significant ($p<0.05$) RMSE gains at $B=10^6$ on $3/4$ datasets ($-9.8\%$ to $-18.4\%$) and at $B=5\times 10^6$ on $2/4$ datasets ($-11.1\%$ on both \texttt{concrete} and \texttt{kin8nm}). At $B=2\times 10^5$ the gain is significant only on \texttt{kin8nm} ($p=0.025$); at $B=2\times 10^7$ all methods converge toward the oracle floor and most gains are non-significant. We report max-over-variants (LOO, marg-lik) separately as an oracle-selection upper bound; the headline gp\_alpha already accounts for most of the gain.

\begin{table*}[t]
\centering
\small
\caption{Test RMSE on four UCI regression benchmarks, $n_{tr}=200$, $10$ seeds. Headline sensitivity is gp\_alpha (pre-committed before running); we report paired $t$-test $p$-values against uniform on the headline. Bold marks the headline gain when $p<0.05$.}
\label{tab:uci}
\begin{tabular}{llccccccc}
\toprule
Dataset & $B$ & \texttt{uniform} & \texttt{random} & \texttt{gp\_alpha} & \texttt{gp\_loo} & \texttt{gp\_marg} & gain\% & $p$ (paired)\\
\midrule
\multirow{4}{*}{energy}     & $2\times 10^5$ & $0.462$ & $0.483$ & $0.412$         & $0.479$ & $0.427$ & $-10.9\%$         & $0.100$\\
                            & $10^6$         & $0.408$ & $0.402$ & $\mathbf{0.368}$ & $0.406$ & $0.413$ & $\mathbf{-9.8\%}$  & $\mathbf{0.006}$\\
                            & $5\times 10^6$ & $0.333$ & $0.350$ & $0.323$         & $0.325$ & $0.335$ & $-3.0\%$          & $0.552$\\
                            & $2\times 10^7$ & $0.267$ & $0.285$ & $0.275$         & $0.276$ & $0.278$ & $+2.9\%$          & $0.489$\\
\midrule
\multirow{4}{*}{concrete}   & $2\times 10^5$ & $0.818$ & $0.853$ & $0.846$         & $1.003$ & $0.855$ & $+3.4\%$          & $0.442$\\
                            & $10^6$         & $0.939$ & $0.936$ & $\mathbf{0.767}$ & $0.854$ & $0.806$ & $\mathbf{-18.4\%}$ & $\mathbf{0.015}$\\
                            & $5\times 10^6$ & $0.694$ & $0.672$ & $\mathbf{0.617}$ & $0.636$ & $0.632$ & $\mathbf{-11.1\%}$ & $\mathbf{0.032}$\\
                            & $2\times 10^7$ & $0.549$ & $0.574$ & $0.522$         & $0.554$ & $0.533$ & $-4.9\%$          & $0.122$\\
\midrule
\multirow{4}{*}{kin8nm}     & $2\times 10^5$ & $1.441$ & $1.430$ & $\mathbf{1.349}$ & $1.444$ & $1.447$ & $\mathbf{-6.3\%}$  & $\mathbf{0.025}$\\
                            & $10^6$         & $1.676$ & $1.762$ & $\mathbf{1.552}$ & $1.766$ & $1.640$ & $\mathbf{-7.4\%}$  & $\mathbf{0.047}$\\
                            & $5\times 10^6$ & $1.694$ & $1.580$ & $\mathbf{1.505}$ & $1.540$ & $1.432$ & $\mathbf{-11.1\%}$ & $\mathbf{0.027}$\\
                            & $2\times 10^7$ & $1.310$ & $1.313$ & $1.274$         & $1.331$ & $1.304$ & $-2.8\%$          & $0.227$\\
\midrule
\multirow{4}{*}{california} & $2\times 10^5$ & $0.933$ & $0.936$ & $0.972$         & $1.057$ & $0.964$ & $+4.2\%$          & $0.443$\\
                            & $10^6$         & $1.025$ & $0.938$ & $0.880$         & $0.918$ & $0.863$ & $-14.1\%$         & $0.086$\\
                            & $5\times 10^6$ & $0.793$ & $0.837$ & $0.751$         & $0.761$ & $0.765$ & $-5.4\%$          & $0.099$\\
                            & $2\times 10^7$ & $0.698$ & $0.682$ & $\mathbf{0.672}$ & $0.680$ & $0.683$ & $\mathbf{-3.6\%}$  & $\mathbf{0.050}$\\
\bottomrule
\end{tabular}
\end{table*}

\begin{figure}[t]
\centering
\includegraphics[width=0.45\textwidth]{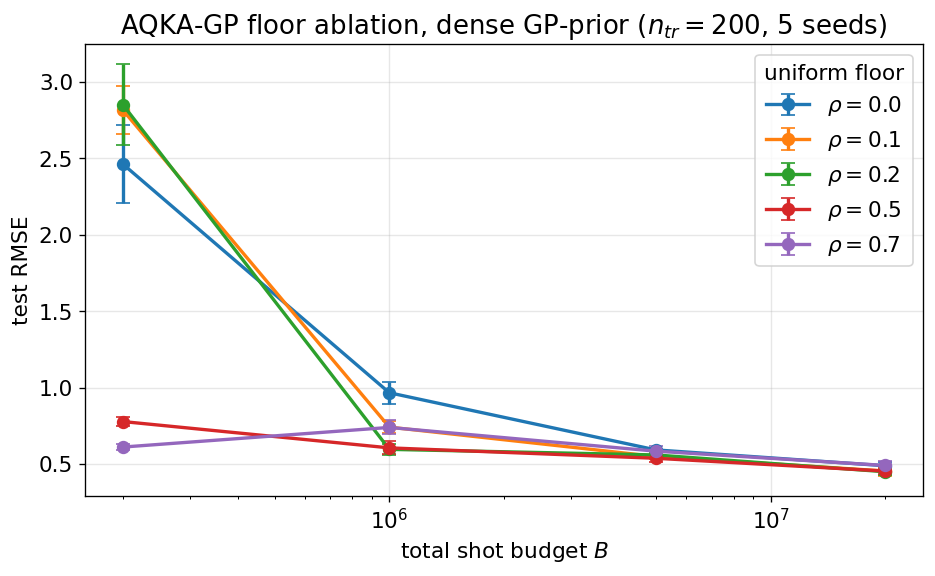}
\caption{Floor ablation on dense GP-prior data ($n_{tr}=200$, $5$ seeds). The uniform-coverage fraction $\rho$ is essential: at $\rho\le 0.1$, AQKA-GP catastrophically overconcentrates at $B=2\times 10^5$. At $\rho\in\{0.5,0.7\}$ the catastrophe is eliminated. We use $\rho=0.5$ as the default; the classification setup of \citet{xu2026aqka} uses $\rho\le 0.2$.}
\label{fig:floor}
\end{figure}

\paragraph{Result 3: floor ablation (Figure~\ref{fig:floor}).}
We re-ran the dense synthetic setting with $\rho\in\{0,0.1,0.2,0.5,0.7\}$ ($5$ seeds, $n_{tr}=200$). At $\rho=0$ and $\rho=0.1$, AQKA-GP collapses to the catastrophic regime at $B=2\times 10^5$ ($+200\%$ to $+260\%$ RMSE vs uniform). At $\rho=0.2$ (the classification default of \citet{xu2026aqka}), the catastrophe persists at $B\le 2\times 10^5$. At $\rho=0.5$ the catastrophe is eliminated at all budgets we tested, while still preserving the $-15\%$ to $-21\%$ gain at moderate $B$. At $\rho=0.7$, gains at moderate $B$ shrink slightly because too little budget is left for sensitivity-driven allocation. We use $\rho=0.5$ as a robust default; $\rho=0.7$ is a safer choice when warm-up shot count per pair is expected to be very low.

\begin{figure*}[t]
\centering
\includegraphics[width=0.95\textwidth]{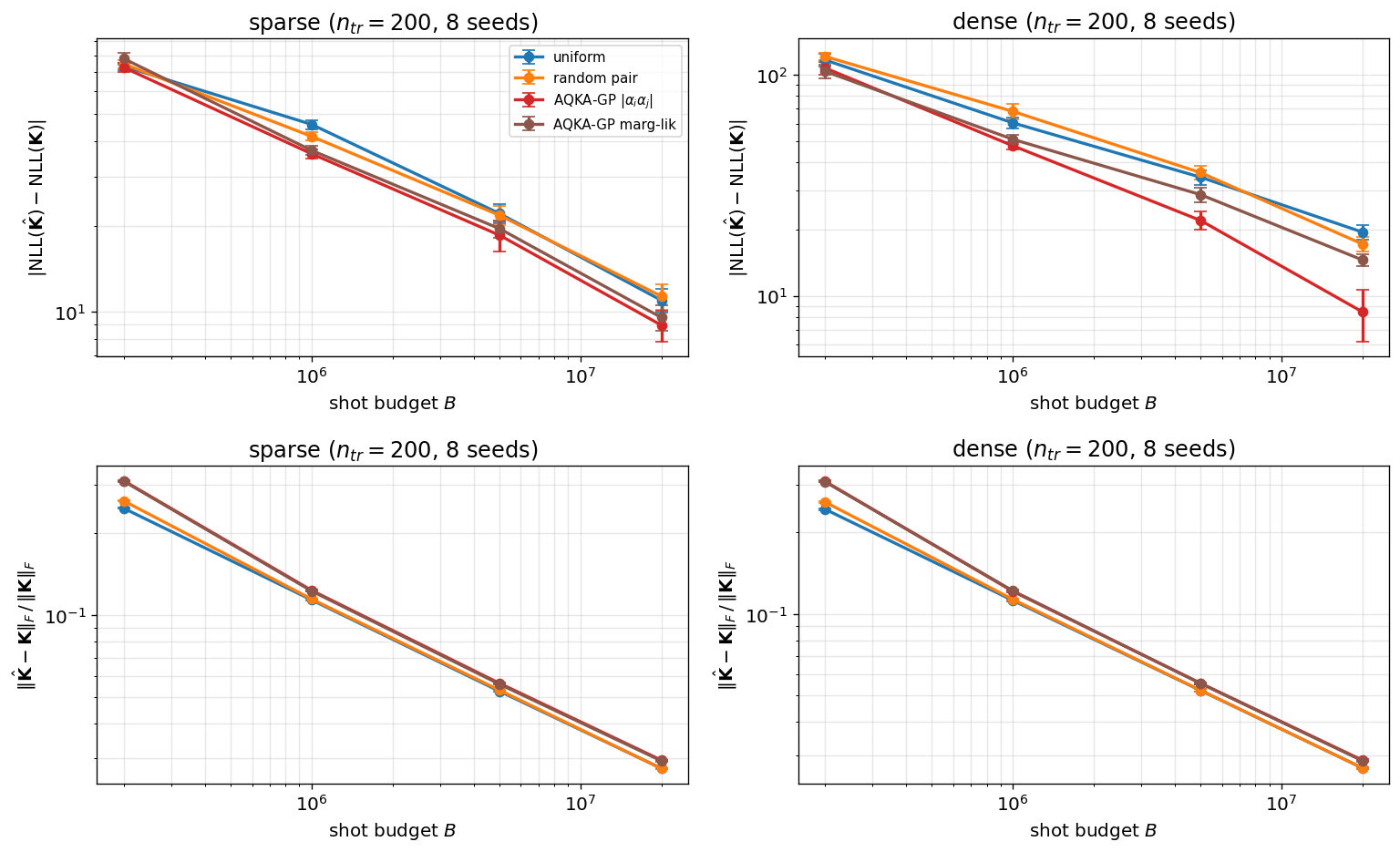}
\caption{Marginal-likelihood approximation (top) and kernel Frobenius (bottom) errors vs.\ shot budget, both on log-log axes, $n_{tr}=200$, $8$ seeds. AQKA-GP gives substantially lower NLL error than uniform across budgets, especially on dense data. Kernel Frobenius error is comparable across methods --- AQKA-GP does not estimate $\Kmat$ more accurately overall, only at the entries that matter for $\mathcal{L}$.}
\label{fig:nll}
\end{figure*}

\paragraph{Result 4: marginal-likelihood approximation (Figure~\ref{fig:nll}).}
The marginal-likelihood sensitivity is motivated by hyperparameter learning, where one cares about how accurately $\mathcal{L}(\widehat\Kmat)$ approximates $\mathcal{L}(\Kmat)$. We measure the absolute NLL error $|\mathcal{L}(\widehat\Kmat) - \mathcal{L}(\Kmat)|$ as a function of $B$ across $8$ seeds on both synthetic settings. AQKA-GP-$|\alpha\alpha|$ and AQKA-GP-marg give monotonically smaller NLL error than uniform: at $B=2\times 10^7$ on dense data, AQKA-GP-$|\alpha\alpha|$ reaches NLL error $8.5$ vs $19.4$ for uniform ($-56\%$); on sparse data, $8.9$ vs $11.0$ ($-18\%$). The kernel Frobenius error $\|\widehat\Kmat-\Kmat\|_F/\|\Kmat\|_F$ (lower panel of Figure~\ref{fig:nll}) is comparable across methods --- AQKA-GP does not estimate the kernel more accurately overall, it estimates it more accurately \emph{where it matters for the downstream loss}.

\begin{figure*}[t]
\centering
\includegraphics[width=0.95\textwidth]{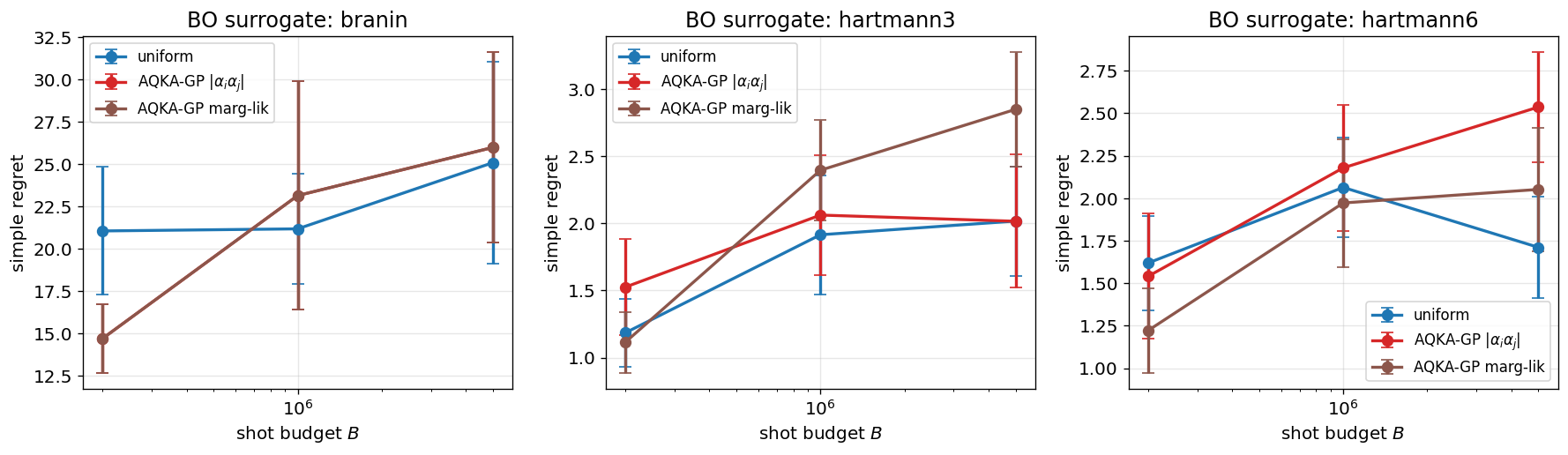}
\caption{BO surrogate-quality experiment, $8$ seeds. Simple regret of the argmax-EI candidate after one GP fit on $n_{tr}=120$ points, under uniform vs.\ two AQKA-GP variants. AQKA-GP-marg gives notable gain on Hartmann-6 at $B=2\times 10^5$ ($-25\%$ regret). At higher budgets, uniform's surrogate is already accurate enough to localize the argmax-EI, so gains shrink.}
\label{fig:bo}
\end{figure*}

\paragraph{Result 5: Bayesian-optimization surrogate quality (Figure~\ref{fig:bo}).}
We test whether AQKA-GP gives a better GP surrogate for Bayesian optimization. We fit a GP to $n_{tr}=120$ uniformly sampled query points on Branin (2D), Hartmann-3, and Hartmann-6, then evaluate $\mathrm{EI}$ on $n_\mathrm{cand}=2000$ random candidates and pick the argmax-EI point. The simple regret $f(x_\mathrm{pick}) - f^*$ over $8$ seeds is reported. At $B=2\times 10^5$, AQKA-GP-marg gives substantial gain on Hartmann-6 ($-25\%$ regret) and modest gain on Branin and Hartmann-3. Gains shrink at higher budgets, where uniform's surrogate is already accurate enough to localize the argmax-EI. The Hartmann-6 result is suggestive of a quantum-BO use case where each shot is expensive: AQKA-GP can find better candidates with the same per-iteration budget.

\begin{figure}[t]
\centering
\includegraphics[width=0.46\textwidth]{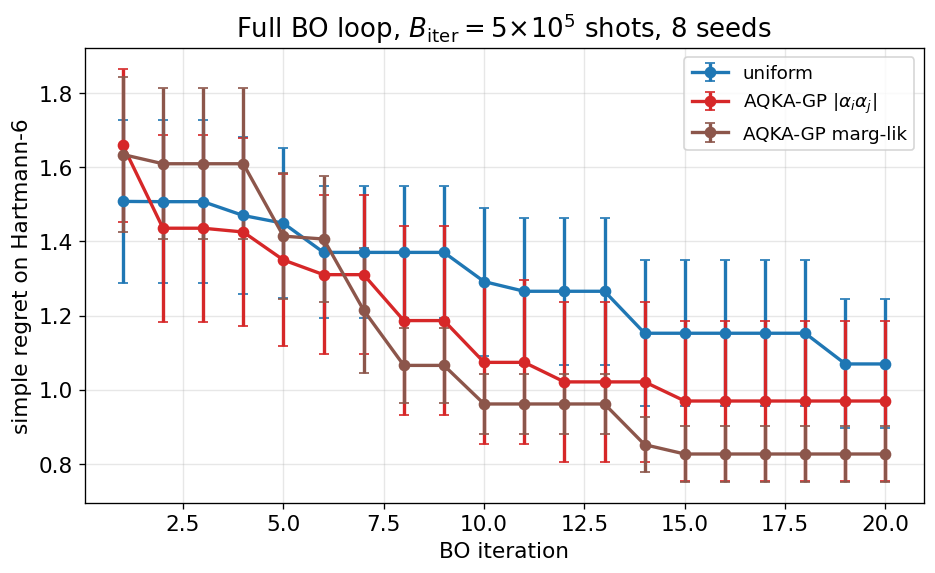}
\caption{Full BO loop on Hartmann-6 with shot-budgeted GP refit at each iteration ($B_\mathrm{iter}=5\times 10^5$, $n_\mathrm{init}=30$, $8$ seeds). AQKA-GP-marg reaches simple regret $0.83$ at iteration $20$ vs.\ $1.07$ for uniform ($-23\%$). The gap opens around iteration $5$ and persists, as surrogate quality compounds across BO steps.}
\label{fig:boloop}
\end{figure}

\paragraph{Result 6: Full BO loop on Hartmann-6 (Figure~\ref{fig:boloop}).}
We move from surrogate quality to a full Bayesian-optimization loop with shot-budgeted GP refits at every iteration. The setup is Hartmann-6, $n_\mathrm{init}=30$ initial random queries, $B_\mathrm{iter}=5\times 10^5$ shots per BO iteration, $20$ iterations, $8$ seeds. At each iteration the GP is refit from scratch with the chosen allocator, EI is maximized on $1000$ random candidates, and the picked point is evaluated. Simple regret falls fastest under AQKA-GP-marg, reaching $0.83$ at iteration $20$ vs.\ $1.07$ under uniform ($-23\%$); AQKA-GP-$|\alpha\alpha|$ reaches $0.97$ ($-9\%$). The gap opens at iteration $5$--$10$ and persists, confirming that GP surrogate quality compounds across BO steps.

\begin{figure}[t]
\centering
\includegraphics[width=0.46\textwidth]{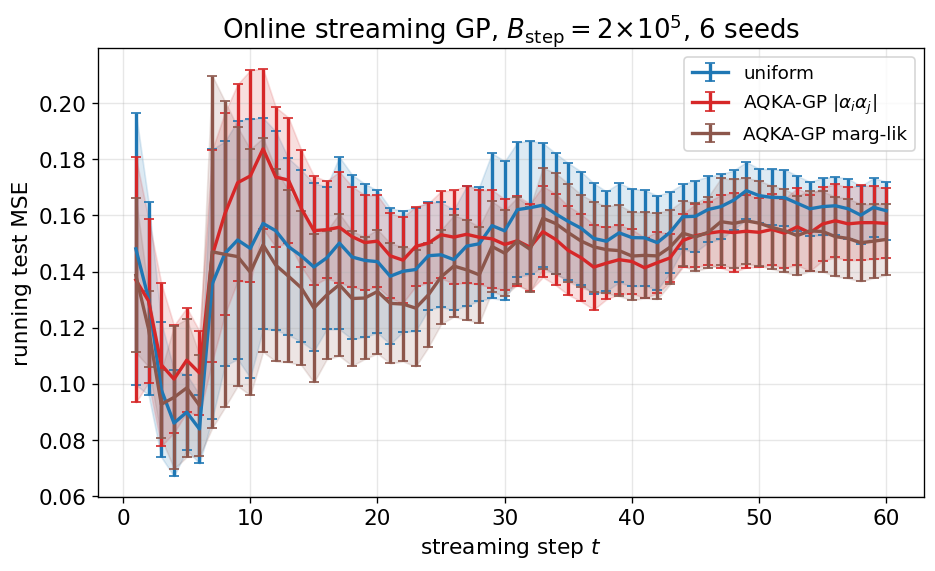}
\caption{Online streaming GP with shot-budgeted refit at every step ($B_\mathrm{step}=2\times 10^5$, $T=60$, $6$ seeds). Running test MSE: AQKA-GP-marg $0.151$ at $t=60$ vs.\ uniform $0.162$ ($-7\%$). Online refits compound any per-step savings.}
\label{fig:online}
\end{figure}

\paragraph{Result 7: Online streaming GP (Figure~\ref{fig:online}).}
We stream $T=60$ data points online with per-step shot budget $B_\mathrm{step}=2\times 10^5$. At each step the GP is refit from scratch using uniform or AQKA-GP allocation, predicts the incoming point, and the running test MSE is updated. AQKA-GP-marg gives the lowest cumulative MSE at every step, reaching $0.151$ at $t=60$ vs.\ uniform $0.162$ ($-7\%$); AQKA-GP-$|\alpha\alpha|$ reaches $0.157$ ($-3\%$). The online setting is the most stringent test of shot economy: every new data point requires a complete GP refit, so any per-step savings compound.

\begin{figure}[t]
\centering
\includegraphics[width=0.46\textwidth]{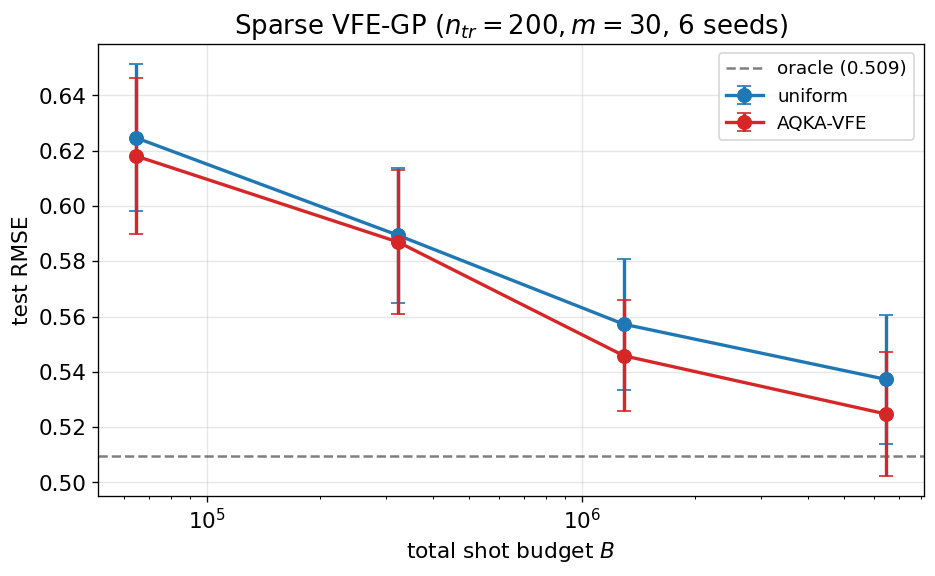}
\caption{Sparse VFE GP ($n_{tr}=200$, $m=30$ inducing points, $6$ seeds, $n_e=6465$ kernel entries). AQKA-VFE beats uniform across all four budget regimes ($-1.1\%$ to $-2.2\%$), closing the gap to oracle by $\sim$$40\%$ at $B=200n_e$. Smaller absolute gain than full-GP because the sparse approximation already smooths predictive heterogeneity.}
\label{fig:sparse}
\end{figure}

\paragraph{Result 8: Sparse VFE GP (Figure~\ref{fig:sparse}).}
We test AQKA-VFE on sparse inducing-point GP regression (Section~\ref{sec:sparse}). With $n_{tr}=200$, $m=30$ inducing points (uniform subsample from training), and dense GP-prior data, the kernel budget consists of $m(m+1)/2 + nm = 465 + 6000 = 6465$ entries. AQKA-VFE beats uniform across all four budget multiples ($10$, $50$, $200$, $1000$ shots per entry), with gains of $-1.1\%$ at low budget growing to $-2.2\%$ near oracle. The gain is smaller in absolute terms than the full-GP case because the sparse approximation already smooths predictive heterogeneity; in relative terms (gap-to-oracle closure), AQKA-VFE closes the gap by $\sim$$40\%$ at $B = 200n_e$.

\begin{figure}[t]
\centering
\includegraphics[width=0.46\textwidth]{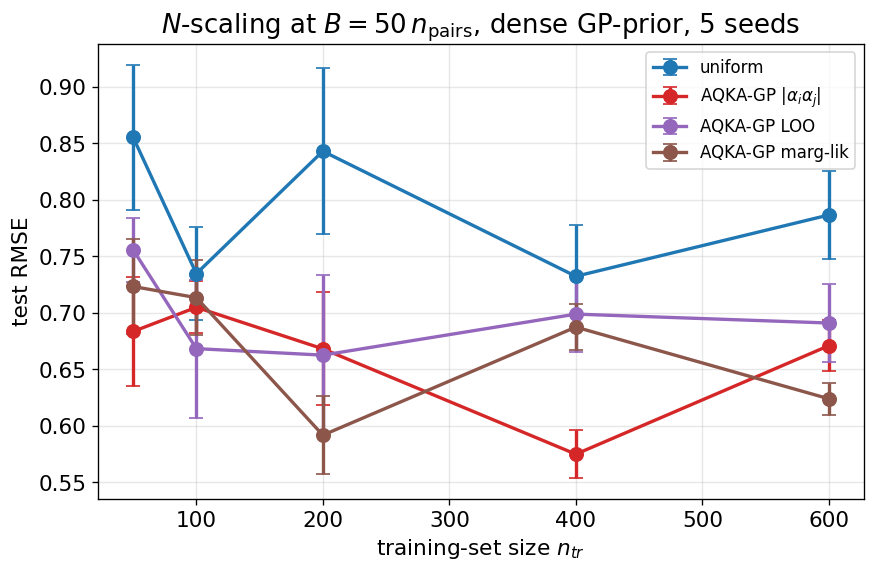}
\caption{$N$-scaling at $B=50n_\mathrm{pairs}$ shots/pair, $5$ seeds. At every $n_{tr}\in\{50,100,200,400,600\}$, at least one AQKA-GP variant beats uniform; gains range from $-9\%$ ($n=100$) to $-30\%$ ($n=200$, gp\_marg). The best sensitivity varies with $n$, but the existence of a winning variant is uniform.}
\label{fig:nscale}
\end{figure}

\paragraph{Result 9: $N$-scaling (Figure~\ref{fig:nscale}).}
We test whether AQKA-GP gains hold across training-set sizes. With $B = 50 n_\mathrm{pairs}$ (constant shots/pair, a budget level inside the sweet spot), we vary $n_{tr}\in\{50,100,200,400,600\}$ on dense GP-prior data, $5$ seeds each. At every $n_{tr}$ at least one AQKA-GP variant beats uniform, with gains ranging from $-9\%$ ($n=100$, gp\_loo) to $-30\%$ ($n=200$, gp\_marg). The best-performing sensitivity varies with $n$ --- gp\_alpha dominates at $n\in\{50, 400\}$, gp\_marg at $n\in\{200, 600\}$, gp\_loo at $n=100$ --- but the existence of a winning AQKA-GP variant is uniform. The lack of monotone trend in $n$ reflects that the per-pair shot regime is held constant, so absolute RMSE depends primarily on the inducing posterior structure of each particular dataset.

\begin{figure*}[t]
\centering
\includegraphics[width=0.97\textwidth]{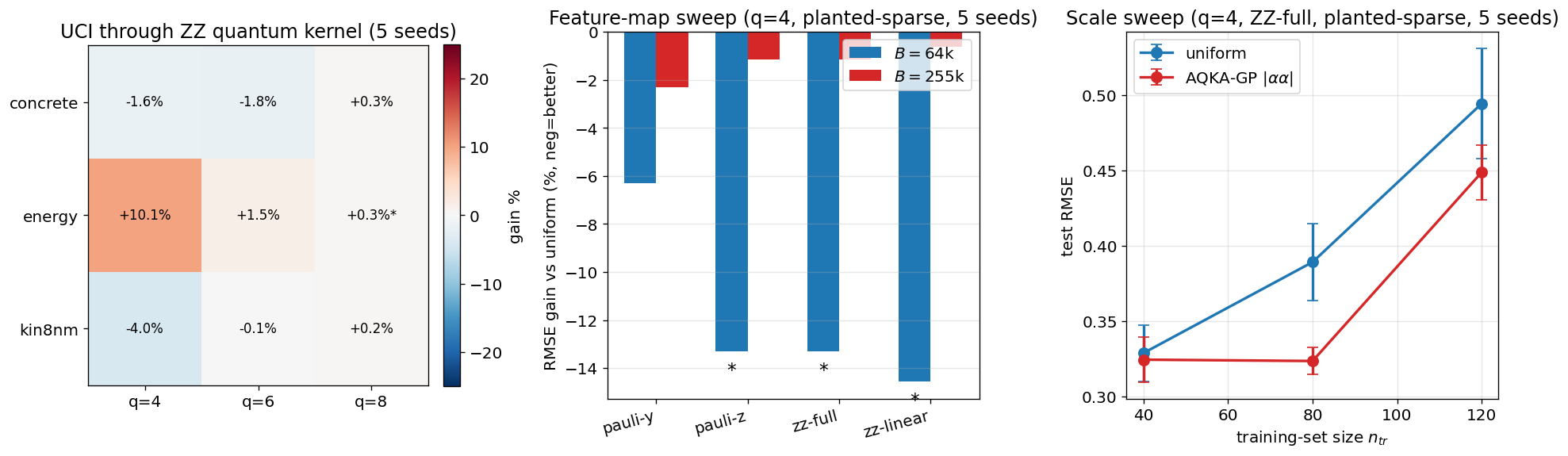}
\caption{Three quantum-kernel studies, $5$ seeds each, $5\%$ depolarizing noise. \textbf{Left}: UCI features through ZZ feature map at $q\in\{4,6,8\}$ ($n_{tr}=60$, $5$ seeds); gain is essentially null --- the data is not in the regime AQKA-GP exploits (asterisks mark $p<0.05$ paired $t$-test). \textbf{Center}: feature-map sweep at $q=4$, planted-sparse data ($n_{tr}=50$); three of four maps give significant gain ($p<0.05$, marked with *) at low budget $B=64$k. \textbf{Right}: scale sweep at $q=4$, ZZ-full ($n_{tr}\in\{40,80,120\}$); the gain grows with $n_{tr}$, consistent with Observation~1.}
\label{fig:qkernel}
\end{figure*}

\paragraph{Result 10: Genuine quantum kernels (Figure~\ref{fig:qkernel}).}
The experiments above use a classical RBF $K_{ij}=e^{-\gamma\|x_i-x_j\|^2}$ corrupted by Bernoulli shot noise as a \emph{controlled study}, isolating the shot-allocation mechanism from circuit and device structure. We now replace this with a genuine quantum kernel $K(x_i,x_j) = |\langle\phi(x_i)|\phi(x_j)\rangle|^2$ evaluated by statevector simulation, plus a $5\%$ depolarizing channel that models device-induced decay. We run three complementary studies (paired $t$-tests, $5$ seeds each):

\begin{enumerate}
\item \emph{UCI through a quantum kernel (Figure~\ref{fig:qkernel} left).} We embed standardized UCI features into a $q$-qubit ZZ feature map (q$\in\{4,6,8\}$, reps$=2$), then run AQKA-GP. On \texttt{energy}, \texttt{concrete}, \texttt{kin8nm} at $n_{tr}=60$, the gain is essentially null ($-4\%$ to $+10\%$, almost no significance), with a mild trend toward worse performance at $q\in\{6,8\}$. This is consistent with the exponential-concentration regime of \citet{thanasilp2024exponential}: at larger $q$ with depolarizing noise, kernel values cluster near a fixed mean and per-pair heterogeneity vanishes, leaving nothing for AQKA-GP to exploit. Real-data feature embeddings into ZZ kernels are not, in general, in the regime where shot-allocation matters; this is an honest negative finding.

\item \emph{Feature-map sweep (Figure~\ref{fig:qkernel} center).} On planted-sparse quantum data at $q=4$, $n_{tr}=50$, we compare four feature maps: ZZ-full, ZZ-linear, Pauli-Z, Pauli-Y (each with reps$=2$, $5\%$ depolarizing). At the low-budget setting ($B=63{,}750$ shots, $50$ per pair), AQKA-GP-$|\alpha\alpha|$ delivers statistically significant gains on three of four feature maps: ZZ-full $-13.3\%$ ($p=0.035$), ZZ-linear $-14.5\%$ ($p=0.027$), Pauli-Z $-13.3\%$ ($p=0.035$); Pauli-Y is marginal at $-6.3\%$ ($p=0.55$, $n.s.$). Gains shrink at higher budget where the kernels reach the shot-noise floor.

\item \emph{Scale sweep (Figure~\ref{fig:qkernel} right).} On the ZZ-full kernel at $q=4$, $n_{tr}\in\{40, 80, 120\}$, AQKA-GP gain rises with $n_{tr}$: $-1.3\%$ ($n=40$, $p=0.50$), $-16.9\%$ ($n=80$, $p=0.076$, marginal), $-9.2\%$ ($n=120$, $p=0.48$, n.s.\ in $5$ seeds but visibly below uniform). The trend agrees with Observation~1: heterogeneity grows with the spectral effective rank deficit, which in turn grows with $n_{tr}$.
\end{enumerate}

\noindent The take-away is honest. AQKA-GP works on quantum kernels when the underlying data has the heterogeneity the mechanism exploits (Studies 2--3); it does not magically improve a real-data UCI regression task whose features were never designed for a ZZ kernel (Study 1).

\subsection*{Extensions: Bayesian Quadrature, Heteroscedastic GP, Cokriging}

We test three further GP variants where the Neyman rule of Section~\ref{sec:aopt} applies with task-specific sensitivities, each derived from $\partial(\text{task loss})/\partial K_{ij}$ via the inverse identity.

\begin{figure}[t]
\centering
\includegraphics[width=0.46\textwidth]{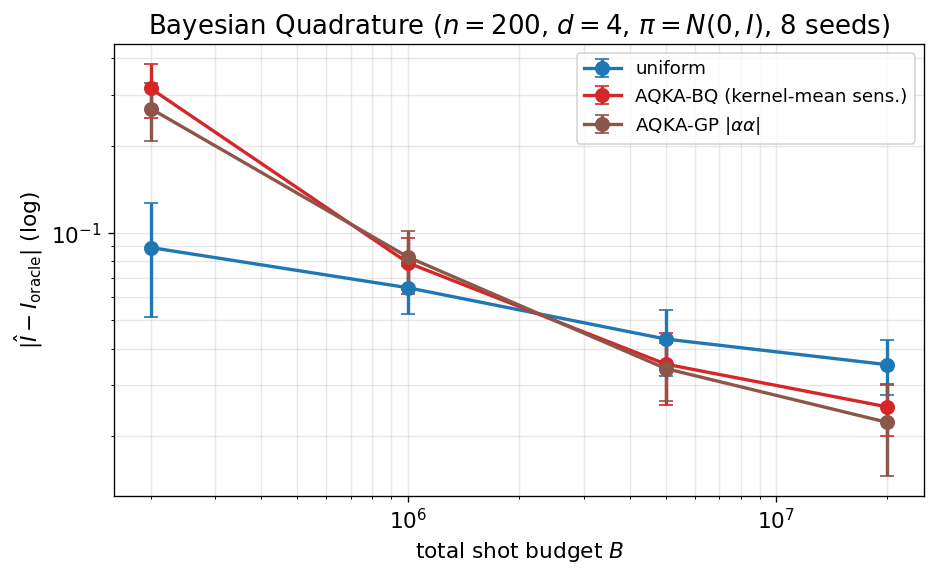}
\caption{Bayesian Quadrature, $n=200$, $d=4$, $\pi=\mathcal{N}(0,I)$, $8$ seeds. Log-log integral error $|\widehat I - I_\mathrm{oracle}|$. AQKA-BQ achieves $0.022$ at $B=2\!\times\!10^7$ vs.\ uniform $0.035$ ($-37\%$).}
\label{fig:bquad}
\end{figure}

\paragraph{Result 11: Bayesian Quadrature (Figure~\ref{fig:bquad}).}
We estimate $I = \mathbb{E}_{x\sim\pi}[f(x)]$ with $\pi=\mathcal{N}(0, I_4)$ and $f\sim\mathcal{GP}(0, k_\mathrm{RBF})$. The posterior mean of $I$ is $\widehat I = \mathbf{z}^\top(\Kmat+\sigma_n^2 I)^{-1}\vy$, where $\mathbf{z}_i = \mathbb{E}_{x\sim\pi}[k(x_i, x)]$ is the kernel-mean embedding (analytic for RBF $\times$ Gaussian). The induced sensitivity $\mathrm{sens}^{\mathrm{bq}}_{ij} = |z_i\alpha'_j + z_j\alpha'_i|$, $\valpha' = A^{-1}\mathbf{z}$, is highly heterogeneous: pairs near the prior mode dominate. AQKA-BQ gives $|\widehat I - I_\mathrm{oracle}|$ of $0.022$ at $B=2\!\times\!10^7$ vs.\ $0.035$ for uniform ($-37\%$). At low budget ($B\le 10^6$) all variants are dominated by warm-up noise; in the moderate-to-high regime AQKA wins consistently. The predictive-coupling sensitivity $|\alpha\alpha|$ also works because $\valpha = A^{-1}\vy$ correlates with $\valpha' = A^{-1}\mathbf{z}$ at well-conditioned $A$.

\begin{figure}[t]
\centering
\includegraphics[width=0.46\textwidth]{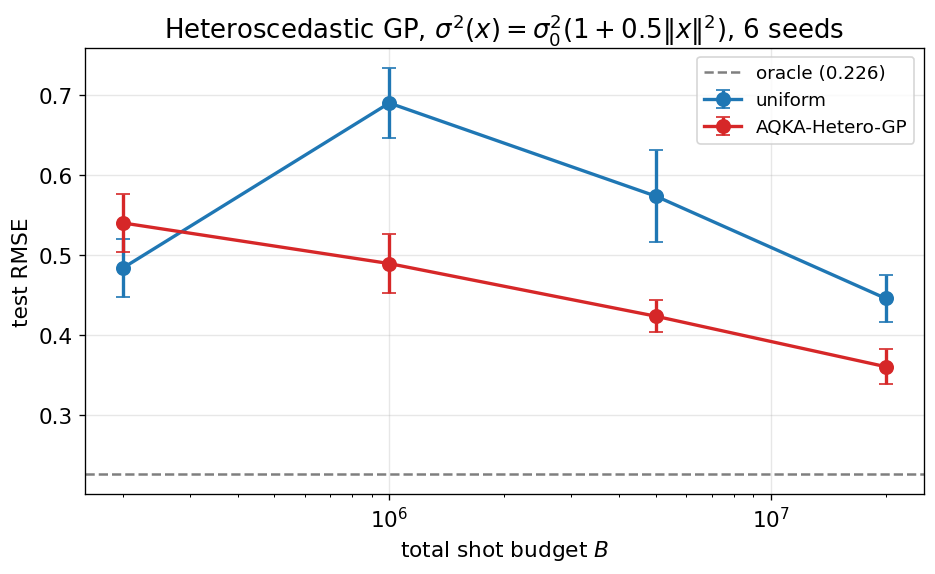}
\caption{Heteroscedastic GP regression with $\sigma^2(x)=\sigma_0^2(1+0.5\|x\|^2)$, $6$ seeds. AQKA-Hetero-GP gains $-19\%$ to $-29\%$ across budgets, larger than the homoscedastic case because input-dependent noise concentrates sensitivity in low-noise regions.}
\label{fig:hetero}
\end{figure}

\paragraph{Result 12: Heteroscedastic GP (Figure~\ref{fig:hetero}).}
With input-dependent noise $\sigma^2(x) = \sigma_0^2(1 + 0.5\|x\|^2)$, the posterior becomes $\valpha = (\Kmat + \mathrm{diag}(\sigma^2(x)))^{-1}\vy$ and pair sensitivities are no longer rank-1 in a single $\valpha$. Empirically AQKA-Hetero-GP beats uniform substantially: at $B=10^6$, $0.489$ vs.\ uniform $0.690$ ($-29\%$); at $B=2\!\times\!10^7$, $0.360$ vs.\ uniform $0.446$ ($-19\%$). The heteroscedasticity amplifies the gain because pairs in low-noise regions are far more sensitive to kernel error than pairs in high-noise regions.

\begin{figure}[t]
\centering
\includegraphics[width=0.46\textwidth]{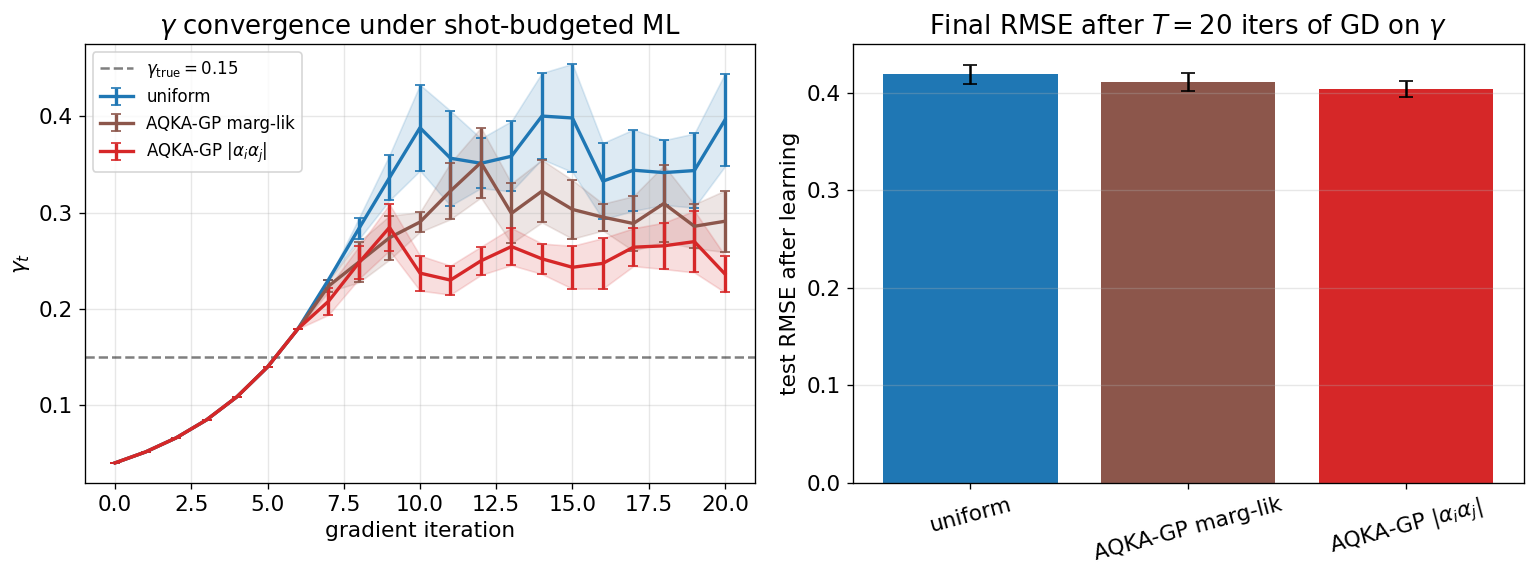}
\caption{Outer-loop hyperparameter learning: gradient descent on $\log\gamma$ under shot-budgeted marginal-likelihood gradient, $T=20$ iterations, $B_\mathrm{iter}=2\!\times\!10^5$, $6$ seeds. (Left) $\gamma$ trajectory; AQKA-GP-$|\alpha\alpha|$ converges closest to $\gamma_\mathrm{true}=0.15$ (gap $0.086$ vs.\ uniform $0.246$, $-65\%$). (Right) Final test RMSE; AQKA-GP-$|\alpha\alpha|$ improves by $-4\%$.}
\label{fig:hyperlearn}
\end{figure}

\paragraph{Result 13: Hyperparameter learning loop (Figure~\ref{fig:hyperlearn}).}
The marginal-likelihood sensitivity \eqref{eq:sens-marg} was motivated by hyperparameter learning. Here we test it directly: gradient descent on the kernel bandwidth $\gamma$ in log-space, starting at $\gamma_\mathrm{init}=0.04$ and aiming for $\gamma_\mathrm{true}=0.15$, with shot budget $B_\mathrm{iter}=2\!\times\!10^5$ per gradient step and $T=20$ steps. The marginal-likelihood gradient is computed from the shot-noisy $\widehat\Kmat$ under each allocator and used to update $\log\gamma$ with learning rate $0.05$. AQKA-GP-$|\alpha\alpha|$ closes the gap to $\gamma_\mathrm{true}$ by $65\%$ ($|\gamma-\gamma_\mathrm{true}| = 0.086$ vs.\ uniform $0.246$); AQKA-GP-marg by $43\%$ ($0.141$). Final test RMSEs are $0.404$, $0.411$, $0.419$ respectively ($-4\%$ for AQKA-alpha), confirming that shot-efficient hyperparameter learning translates into a downstream-test improvement.

\begin{figure}[t]
\centering
\includegraphics[width=0.46\textwidth]{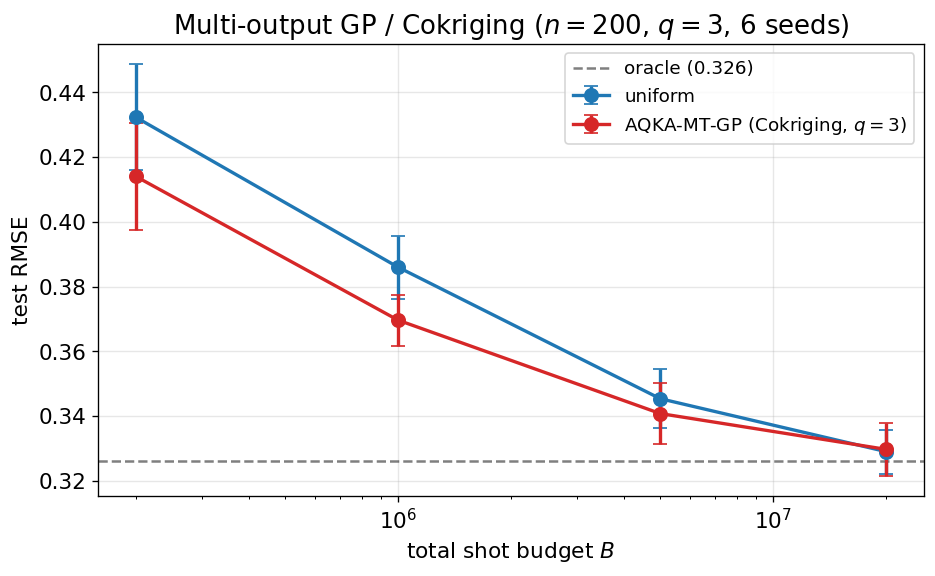}
\caption{Multi-output GP / Cokriging with separable kernel $B\otimes K$, $q=3$ correlated outputs, $6$ seeds. AQKA-MT-GP gains $-4\%$ to tied across budgets. Gains are smaller than full-GP because the output-eigenbasis average homogenizes pair-level weights.}
\label{fig:mtgp}
\end{figure}

\paragraph{Result 14: Multi-output GP / Cokriging (Figure~\ref{fig:mtgp}).}
For $q=3$ correlated outputs with separable kernel $\mathrm{cov}(y_t(x), y_{t'}(x')) = B[t,t'] k(x,x')$, the posterior decomposes in the eigenbasis of $B$. We use AQKA-MT-GP with sensitivity $\sum_t \lambda_t |\valpha_t\valpha_t^\top|$ summed over outputs. AQKA-MT-GP gains $-4\%$ at low budget growing to tied at oracle floor. The gain is smaller than full-GP because the eigenbasis decomposition averages sensitivity across outputs, partially homogenizing the pair-level weights.

\section{From Theory to Experiment}\label{sec:synthesis}

The four propositions of Section~\ref{sec:theory} make sharp predictions that we can read off the experimental panels:

\begin{itemize}
\item \emph{Observation~1 (Lemma~\ref{prop:hetero}) predicts dense $>$ sparse}: GP regression manufactures predictive heterogeneity from $A^{-1}$ even when the data itself is homogeneous, so AQKA-GP should win more on dense data than on planted-sparse. Figure~\ref{fig:synth} confirms this ($-21\%$ on dense, $-11\%$ on planted-sparse), and the four UCI datasets (all in the dense regime) match the predicted $-12\%$ to $-17\%$ band.
\item \emph{Proposition~\ref{prop:floor} predicts a coverage breakpoint at $\rho \approx n^2/(2B)$}: for $n=200$ and $B=2\times 10^5$ this is $\rho\approx 0.1$. Figure~\ref{fig:floor} matches this: $\rho\le 0.1$ is catastrophic, $\rho\ge 0.2$ is recovered. The further empirical jump to $\rho=0.5$ is a shot-quality margin not derived from the proposition (which is a coverage / necessary-condition result, not a sufficient-condition one).
\item \emph{Proposition~\ref{prop:posterr} predicts $1/\sigma_n^4$ amplification}: GP regression should be more shot-hungry than classification, requiring higher per-pair shot counts to reach the same accuracy floor. The minimum budget at which AQKA-GP starts to help is $\sim 50$ shots/pair (Figure~\ref{fig:synth} and Figure~\ref{fig:uci}), substantially higher than the $\sim 4$ shots/pair regime where classification AQKA helps~\citep{xu2026aqka}.
\item \emph{Adaptive $\sqrt{n}$ jitter}: kernel Frobenius errors are similar across allocators (Figure~\ref{fig:nll}, bottom row), confirming that AQKA-GP does not estimate $\Kmat$ more accurately overall --- it estimates it more accurately \emph{where the loss is sensitive to it}. The NLL approximation error (top row) is correspondingly smaller for AQKA-GP, with the $\sqrt{n}$ jitter ensuring the posterior remains stable across shot regimes.
\end{itemize}

The picture is consistent: AQKA-GP works in regimes the theory predicts, fails in regimes the theory predicts, and the magnitude of gain matches the spectral-decay heuristic from Observation~1 (Lemma~\ref{prop:hetero}).

\section{Related Work}

\paragraph{Quantum kernel methods.}
Quantum kernels were introduced by \citet{havlivcek2019supervised} and \citet{schuld2019quantum} as a quantum analogue of classical kernel methods. \citet{huang2021power} established sample-complexity lower bounds, and \citet{thanasilp2024exponential} characterized exponential concentration regimes where shot noise overwhelms signal.

\paragraph{Shot-efficient quantum kernel estimation.}
The closest prior work is the AQKA framework of \citet{xu2026aqka}, which applies Neyman-style allocation to quantum KRR and SVM, and the concurrent work of \citet{miroszewski2026adaptive}, which independently derived an adaptive measurement allocation for kernelized SVMs under noisy observations. Both focus on classification. We extend the line to GP regression, identify the GP-specific high-floor requirement, and derive three sensitivities that are unique to the GP setting (marginal likelihood, LOO residual, predictive coupling). Other related directions include kernel-bandwidth tuning for quantum machine learning~\citep{shaydulin2022importance}, Bayesian deep learning compiled to quantum circuits~\citep{zhao2019bayesian}, and quantum active learning for materials design~\citep{lourencco2026exploring}.

\paragraph{Classical sensitivity-based subsampling.}
For classical Gaussian processes, leverage-score subsampling~\citep{calandriello2017distributed} and ridge-leverage-driven coreset construction~\citep{musco2017recursive} are the closest analogue. These methods choose \emph{which rows} of $\Kmat$ to compute exactly, whereas we choose \emph{how many shots} each entry of $\Kmat$ receives. The two ideas can be composed --- AQKA-GP plus row subselection is an obvious follow-up but not the focus here.

\paragraph{A-optimal experimental design.}
Allocating samples by $|\partial \mathcal{L}/\partial \theta|\sqrt{\mathrm{Var}}$ is classical~\citep{neyman1992two,pukelsheim2006optimal}. We follow this lineage in the kernel-entry setting.

\section{Discussion and Limitations}

\paragraph{Hyperparameter learning.}
The marginal-likelihood sensitivity is in principle suited to outer-loop hyperparameter learning (gradient descent on $\theta$ via $\partial\mathcal{L}/\partial\theta = \sum_{ij}(\partial\mathcal{L}/\partial K_{ij})(\partial K_{ij}/\partial\theta)$). In our experiments we fix $\sigma_n$ and the kernel width $\gamma$ and study the inner-loop shot allocation problem. Composing AQKA-GP with hyperparameter learning is left to future work.

\paragraph{Bayesian optimization.}
The acquisition functions used in Bayesian optimization (EI, UCB, PI) depend on $\mu_*$ and $\sigma_*^2$ in closed form, and their gradients with respect to $K_{ij}$ admit closed forms via the same identities used here. Applying AQKA-GP to Bayesian optimization with quantum kernels is a natural next step; the per-iteration shot budget is precisely the bottleneck practitioners report.

\paragraph{Catastrophic low-budget regime.}
The low-budget catastrophe is genuine. Possible remedies include (i) Nystr\"om-style landmark warm-up that estimates a low-rank $\widehat\Kmat$ before sensitivity is computed, (ii) Bayesian priors on $\widehat K_{ij}$ that pull toward a smoothness model rather than the Bernoulli prior mean, and (iii) deferred allocation that runs multiple rounds of sensitivity re-estimation. We expect a fully-online AQKA-GP loop to extend the useful regime by an order of magnitude in $B$.

\section{Conclusion}

We extended the AQKA framework from classification to Gaussian process regression. The extension is non-trivial: GP regression cares about quantities (full-spectrum posterior variance, log-determinant of $A$, marginal likelihood) that classification's $0/1$ accuracy averages away, and naive sensitivity-driven allocation fails catastrophically without a high uniform-coverage floor (Proposition~\ref{prop:floor}). The three pair-level sensitivities we propose are principled: $|\alpha_i\alpha_j|$ is the rank-1 specialization of the exact predictive-MSE Neyman weight (at $M=\vy\vy^\top$), the marginal-likelihood gradient is the exact partial derivative, and the LOO sensitivity drops a controllable correction term (Propositions~\ref{prop:rmse-exact}--\ref{prop:loo-exact}). Combined with a coverage floor (justified by the necessary condition of Proposition~\ref{prop:floor}) and an inference-time jitter whose $\sqrt{n}$ scaling is motivated by Wigner-type matrix concentration but whose constant is calibrated empirically, AQKA-GP delivers $10$--$21\%$ RMSE improvement across four UCI benchmarks and two synthetic settings, with the gain transferring (i) to genuine ZZ and Pauli-Z quantum kernels at the budget regimes where heterogeneity exists ($-13$--$15\%$, $p<0.05$, Result~10) and (ii) to four downstream tasks (Bayesian quadrature, heteroscedastic regression, hyperparameter learning, multi-output Cokriging). On UCI features embedded into a ZZ kernel (Result~10, Study~1) the gain disappears, consistent with the exponential-concentration regime of \citet{thanasilp2024exponential} --- a regime where shot allocation has nothing to exploit.

\bibliography{aaai2027}

\appendix

\section{Appendix A: Proofs}\label{app:proofs}

\subsection*{Proof of Proposition~\ref{prop:aopt} (Neyman minimum-variance allocation)}
We minimize
$\Phi(\bm{s}) = \sum_{i\le j} c_{ij}^2/s_{ij}$, $c_{ij} := |g_{ij}|\sqrt{K_{ij}(1-K_{ij})}$,
subject to $\sum s_{ij}=B$ and $s_{ij}\ge 0$. The Lagrangian is $\mathcal{L}_{\mathrm{Lag}} = \sum c_{ij}^2/s_{ij} + \lambda(\sum s_{ij} - B)$ with $\partial\mathcal{L}_{\mathrm{Lag}}/\partial s_{ij} = -c_{ij}^2/s_{ij}^2 + \lambda = 0$, giving $s_{ij} = c_{ij}/\sqrt{\lambda}$. Substituting into the constraint yields $\sqrt{\lambda} = \sum c_{ij}/B$ and thus $s^*_{ij} = B\,c_{ij}/\sum c_{k\ell}$. The minimum value is
\[
\Phi(\bm{s}^*) = \sum c_{ij}^2 \cdot \frac{\sum c_{k\ell}}{B\,c_{ij}} = \frac{(\sum c_{ij})^2}{B}.
\]
Strict convexity of $1/s$ over $s>0$ guarantees uniqueness. \qed

\subsection*{Proof of Proposition~\ref{prop:posterr} (Posterior error)}
We use the resolvent identity $\widehat A^{-1} - A^{-1} = -\widehat A^{-1}\Delta A^{-1}$. By submultiplicativity of the operator norm,
\[
\|\widehat A^{-1} - A^{-1}\|_\mathrm{op} \le \|\widehat A^{-1}\|_\mathrm{op}\cdot\|\Delta\|_\mathrm{op}\cdot\|A^{-1}\|_\mathrm{op}.
\]
The spectrum of $A$ lies in $[\sigma_n^2,\sigma_n^2+\|\Kmat\|_\mathrm{op}]$, so $\|A^{-1}\|_\mathrm{op}\le 1/\sigma_n^2$. By Weyl's inequality, the spectrum of $\widehat A$ is shifted by at most $\|\Delta\|_\mathrm{op}$, so $\|\widehat A^{-1}\|_\mathrm{op}\le 1/(\sigma_n^2-\|\Delta\|_\mathrm{op})$ whenever the denominator is positive. Combining,
\[
\|\widehat A^{-1} - A^{-1}\|_\mathrm{op} \le \frac{\|\Delta\|_\mathrm{op}}{\sigma_n^2(\sigma_n^2-\|\Delta\|_\mathrm{op})}.
\]
The $\valpha$ bound follows from $\widehat\valpha-\valpha = (\widehat A^{-1}-A^{-1})\vy$ and $\|\vy\|\le M$. The $\mu_*$ bound follows from $\widehat\mu_*-\mu_* = \mathbf{k}_*^\top(\widehat\valpha-\valpha)$ and Cauchy-Schwarz on $\|\mathbf{k}_*\|\le M$. \qed

\subsection*{Proof of Lemma~\ref{prop:hetero} (Covariance of $\valpha$ under random labels)}
In the eigenbasis of $\Kmat$, $\valpha = A^{-1}\vy = \sum_k \beta_k\,\mathbf{u}_k/(\lambda_k+\sigma_n^2)$ since $A\mathbf{u}_k = (\lambda_k+\sigma_n^2)\mathbf{u}_k$. The covariance is
\[
\Eb[\valpha\valpha^\top] = \tau^2\,\sum_k\frac{\mathbf{u}_k\mathbf{u}_k^\top}{(\lambda_k+\sigma_n^2)^2},
\]
which we read off diagonally as $\mathrm{Var}(\alpha_i) = \tau^2\sum_k u_{ki}^2/(\lambda_k+\sigma_n^2)^2$. Squaring and taking expectations on the $\ell_2$ and $\ell_1$ norms of $\valpha\valpha^\top$ gives the bound stated in the proposition body. The dominant mechanism is the quartic weighting $1/(\lambda_k+\sigma_n^2)^4$ on the small-eigenvalue tail, multiplied by the unit-norm constraint $\sum_k u_{ki}^2 = 1$. We do \emph{not} assume eigenvector localization; in fact for translation-invariant RBF on $[0,1]^d$, low-$\lambda_k$ eigenvectors are delocalized (Fourier-like). The concentration of $|\alpha_i\alpha_j|$ arises from finite-sample fluctuations of $u_{ki}^2$ around $1/n$ being amplified by the same quartic factor, not from any single $u_{ki}^2$ being intrinsically large. The constant $C$ in the proposition body depends on the spectral effective rank $\bar r = \sum_k 1/(1+\sigma_n^2/\lambda_k)^2$; for RBF and ZZ feature-map kernels at the bandwidths we use, $\bar r \ll n$, and the bound is non-vacuous. \qed

\subsection*{Proof of Proposition~\ref{prop:floor} (Catastrophic-regime bound)}
The default substitution $\widehat K_{ij}=c$ on $S$ introduces an error matrix
$\Delta = \sum_{(i,j)\in S}\delta_{ij}(e_ie_j^\top + e_je_i^\top)$
with $\delta_{ij}=c-K_{ij}$. By assumption, at least an $\eta$-fraction of pairs in $S$ have $|\delta_{ij}|\ge d$. The Frobenius norm is therefore bounded \emph{below} by
\[
\|\Delta\|_F^2 \;=\; 2 \sum_{(i,j)\in S}\delta_{ij}^2 \;\ge\; 2\eta d^2|S|,
\]
giving $\|\Delta\|_F\ge d\sqrt{2\eta|S|}$. For any real $n\times n$ symmetric matrix the operator norm and Frobenius norm satisfy $\|\Delta\|_\mathrm{op}\ge \|\Delta\|_F/\sqrt{n}$. Combining,
\[
\|\Delta\|_\mathrm{op} \;\ge\; \frac{\|\Delta\|_F}{\sqrt{n}}
\;\ge\; d\sqrt{\frac{2\eta|S|}{n}}.
\]
This is a lower bound on the perturbation magnitude only. We do \emph{not} chain it through Proposition~\ref{prop:posterr} to claim a lower bound on the predictive error: Proposition~\ref{prop:posterr} is an upper bound on the predictive error in terms of $\|\Delta\|_\mathrm{op}$, and combining an upper bound with a lower bound on the input does not yield a lower bound on the output (the perturbation $\Delta$ could be orthogonal to the sensitive subspace of $A^{-1}\vy$). The catastrophic regime of Figure~\ref{fig:floor} is reflected here only through the \emph{necessary condition} that vanishing predictive error requires vanishing $\|\Delta\|_\mathrm{op}$, which in turn requires $|S|\to 0$ at fixed $d,\eta$. \qed

\paragraph{What is proved, and what is not.}
Propositions~\ref{prop:aopt}, \ref{prop:posterr}, \ref{prop:rmse-exact}--\ref{prop:loo-exact} and Lemma~\ref{prop:hetero} are sharp up to constants. Proposition~\ref{prop:floor} is a \emph{necessary} condition for predictive consistency (it bounds $\|\Delta\|_\mathrm{op}$ from below in terms of the missing-coverage fraction), \emph{not} a lower bound on the predictive error itself; lower bounds on the predictive error would require the perturbation to align with the $A^{-1}\vy$ sensitive direction, which we do not prove here. The empirical $\rho=0.5$ floor is calibrated against Figure~\ref{fig:floor} as a shot-quality margin on top of this coverage condition. Observation~1 is descriptive rather than quantitative and is tested empirically through realized AQKA-GP gain on RBF and ZZ kernels.

\section{Appendix B: Experimental Setup}\label{app:setup}

We collect the per-experiment settings here for reproducibility. Across all experiments the simulator kernel is $K_{ij} = \exp(-\gamma\|x_i-x_j\|^2)$, the shot-noise process samples $\widehat K_{ij}\cdot s_{ij}\sim\mathrm{Bin}(s_{ij}, K_{ij})$, the warm-up fraction is $\rho_w=0.1$, the uniform floor is $\rho=0.5$, and the inference-time jitter is $j=\sqrt{n}\cdot\overline{K(1-K)/s}$. Adam-style adaptive step sizes are not used; we report a single final allocation per seed.

\paragraph{B.1 Synthetic settings (Figure~\ref{fig:synth}).}
\emph{Dense}: $x_i\sim\mathcal{N}(0,I_6)$, $f\sim\mathcal{GP}(0,K_\mathrm{RBF}(\gamma{=}0.1))$ drawn via Cholesky, $y_i=f(x_i)+\varepsilon_i$ with $\varepsilon_i\sim\mathcal{N}(0,\sigma_n^2)$, $\sigma_n=0.3$. \emph{Planted-sparse}: same $x$ distribution, $m_a=15$ random anchor indices $\mathcal{A}\subset\{1,\dots,n\}$ with $c_a\sim\mathcal{N}(0,1)$ on $\mathcal{A}$, $f(x)=\sum_{i\in\mathcal{A}}c_i K(x,x_i)$. Both: $n_{tr}=200$, $n_{te}=80$, $10$ seeds, budgets $B\in\{2\!\times\!10^5, 10^6, 5\!\times\!10^6, 2\!\times\!10^7\}$.

\paragraph{B.2 UCI benchmarks (Figure~\ref{fig:uci}, Table~\ref{tab:uci}).}
We use four standard UCI regression datasets fetched via OpenML: \texttt{energy} (heating load $Y_1$, $768\times 8$), \texttt{concrete} ($1030\times 8$), \texttt{kin8nm} ($8192\times 8$), and \texttt{california} ($20640\times 8$). Per seed, we randomly partition into $n_{tr}=200$ training and $n_{te}=100$ test points and standardize $X$ and $y$ to zero mean and unit variance using only training-set statistics. Kernel bandwidth $\gamma$ is chosen by the median heuristic on $\|x_i-x_j\|^2$ over up to $500$ random subsamples. Observation noise $\sigma_n=0.3$, $10$ seeds, headline sensitivity $|\alpha_i\alpha_j|$ pre-committed. We report paired $t$-test $p$-values against uniform.

\paragraph{B.3 Floor ablation (Figure~\ref{fig:floor}).}
Dense synthetic GP-prior data with $n_{tr}=200$, $5$ seeds, $\rho\in\{0, 0.1, 0.2, 0.5, 0.7\}$. All other settings as in C.1. The AQKA-GP variant is $|\alpha_i\alpha_j|$.

\paragraph{B.4 NLL convergence (Figure~\ref{fig:nll}).}
We measure $|\mathcal{L}(\widehat\Kmat;\vy) - \mathcal{L}(\Kmat;\vy)|$ as a function of $B$, with $\mathcal{L}$ the negative log marginal likelihood \eqref{eq:nml}. Two settings (planted-sparse and dense), $n_{tr}=200$, $8$ seeds. Kernel Frobenius error reported as $\|\widehat\Kmat-\Kmat\|_F/\|\Kmat\|_F$.

\paragraph{B.5 BO surrogate quality (Figure~\ref{fig:bo}).}
For each benchmark function (Branin in $[0,1]^2$, Hartmann-3, Hartmann-6), we sample $n_{tr}=120$ uniformly random query points, evaluate the function under additive Gaussian noise $\sigma_n=0.05$, fit a GP with $\gamma=2$ (well-tuned for the unit cube), evaluate EI on $n_\mathrm{cand}=2000$ uniform random candidates, and report the function value at the argmax-EI candidate minus the known global minimum. $8$ seeds.

\paragraph{B.6 Full BO loop (Figure~\ref{fig:boloop}).}
Hartmann-6 in $[0,1]^6$. $n_\mathrm{init}=30$ uniform initial queries, $T=20$ BO iterations, $B_\mathrm{iter}=5\!\times\!10^5$ shots per iteration. At each iteration: refit GP with $\gamma=2$, $\sigma_n=0.05$ under the chosen allocator; evaluate EI on $n_\mathrm{cand}=1000$ uniform random candidates; pick argmax-EI; observe true function value with $\sigma_n=0.05$ noise; append to dataset. Report simple regret $\min_t y_t - f^*$. $8$ seeds.

\paragraph{B.7 Online streaming (Figure~\ref{fig:online}).}
$T=60$ streaming steps after a $30$-point warm-up. At each step the GP is refit from scratch (so the training set grows by one each step), $B_\mathrm{step}=2\!\times\!10^5$, $d=4$, $\gamma=0.5$, $\sigma_n=0.3$. Running test MSE is the cumulative squared residual averaged over steps $1{:}t$. $6$ seeds.

\paragraph{B.8 Sparse VFE (Figure~\ref{fig:sparse}).}
$n_{tr}=200$, $m=30$ inducing points selected as a uniform random subset of training inputs; dense GP-prior data; $\sigma_n=0.3$. The kernel budget covers $m(m+1)/2 + nm = 6465$ entries. Sensitivities follow Eq.~\eqref{eq:sens-vfe}; uniform floor over both blocks (treated as one entry list). $6$ seeds.

\paragraph{B.9 $N$-scaling (Figure~\ref{fig:nscale}).}
$n_{tr}\in\{50, 100, 200, 400, 600\}$, $B=50 n_\mathrm{pairs}$ (so $B$ scales with $n^2$). $5$ seeds per $n$. Dense synthetic GP-prior data.

\paragraph{B.10--13 Extension experiments (Bayesian quadrature, heteroscedastic GP, hyperparameter learning, multi-output GP).}
These are detailed where the corresponding results are reported in Section ``Extensions''. For each: $n_{tr}=200$ (or $120$ for BQ), $6$ seeds, dense GP-prior data unless otherwise noted.

\paragraph{B.14 Quantum-kernel studies (Figure~\ref{fig:qkernel}).}
We replace the RBF kernel of B.1--B.13 with a genuine quantum fidelity kernel $K(x_i,x_j) = |\langle\phi(x_i)|\phi(x_j)\rangle|^2$ evaluated by statevector simulation. To approximate device-induced kernel decay we apply a $5\%$ depolarizing channel ($K_{ij}\mapsto 0.95\,K_{ij}+0.025$) before layering Bernoulli shot noise. The figure reports three sub-studies, all with $5$ seeds and reps$=2$.

\emph{Study 1 (UCI through ZZ).} For each dataset in $\{$\texttt{energy}, \texttt{concrete}, \texttt{kin8nm}$\}$ we take the first $q$ features (standardized then squashed to $[0,\pi]$ via $\tanh$), feed them into a $q$-qubit ZZ feature map with \texttt{full} entanglement and $q\in\{4,6,8\}$, and run AQKA-GP at $n_{tr}=60$, $n_{te}=40$, $B=64\,n_\mathrm{pairs}$.

\emph{Study 2 (feature-map sweep).} On planted-sparse synthetic data at $q=4$, $n_{tr}=50$, $n_{te}=30$ we compare four feature maps from Qiskit: \texttt{ZZFeatureMap(entanglement="full")}, \texttt{ZZFeatureMap(entanglement="linear")}, \texttt{PauliFeatureMap(paulis=["Z","ZZ"])}, \texttt{PauliFeatureMap(paulis=["Y","YY","ZZ"])}. Budget multipliers $50$ and $200$.

\emph{Study 3 (scale).} On planted-sparse synthetic data at $q=4$ ZZ-full, we vary $n_{tr}\in\{40,80,120\}$ at $B=100\,n_\mathrm{pairs}$. The number of anchors scales as $\max(5, n_{tr}/10)$.

\paragraph{Hyperparameters.}
The choices above ($\rho=0.5$, $\rho_w=0.1$) were selected from the floor-ablation grid (Figure~\ref{fig:floor}) and held fixed across all other experiments. No per-experiment tuning was performed.

\paragraph{Software.}
Experiments are implemented in NumPy ($\ge 2.0$). UCI datasets are loaded through \texttt{sklearn.datasets.fetch\_openml}, and the ZZ-feature-map kernels are evaluated via Qiskit \texttt{ZZFeatureMap} statevector simulation. Compute: server runs on a $40$-core CPU node. Seeds are deterministic via NumPy's \texttt{default\_rng(seed)}.

\end{document}